\def\paperID{10368}
\def\confName{CVPR}
\def\confYear{2024}

\def\paperTitle{SUGAR \includegraphics[height=15pt]{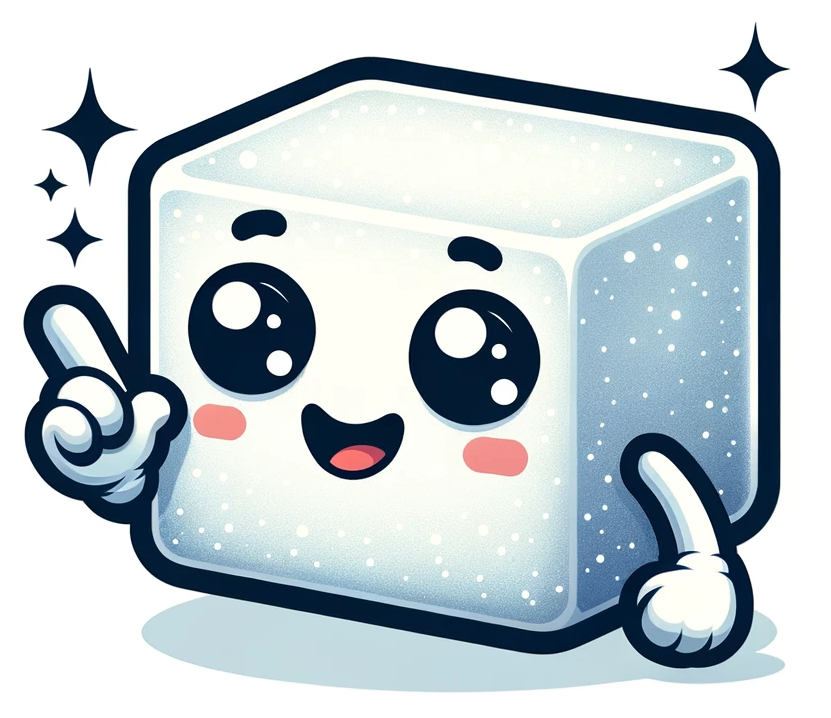}: Pre-training 3D Visual Representations for Robotics}

\def\authorBlock{
    Shizhe Chen$^\dagger$ \qquad
    Ricardo Garcia$^\dagger$ \qquad
    Ivan Laptev$^\star$ \qquad
    Cordelia Schmid$^\dagger$ \\
    {\small $^\dagger$~Inria, \'Ecole normale sup\'erieure, CNRS, PSL Research University} \\
    {\small $^\star$~Mohamed bin Zayed University of Artificial Intelligence} \\
    {\tt\small \url{https://cshizhe.github.io/projects/robot_sugar.html}}
}

\newif\ifreview 
\newif\ifarxiv \newcommand{\arxiv}{\arxivtrue}
\newif\ifcamera 
\newif\ifrebuttal 

\arxiv

\pdfoutput=1
\documentclass[10pt,twocolumn,letterpaper]{article}
\ifreview \usepackage[review]{cvpr} \fi
\ifarxiv \usepackage[pagenumbers]{cvpr} \fi
\ifrebuttal \usepackage[rebuttal]{cvpr} \fi
\ifcamera \usepackage{cvpr} \fi

\usepackage{graphicx}	
\usepackage{amsmath}	
\usepackage{amssymb}	
\usepackage{booktabs}
\usepackage{times}
\usepackage{microtype}
\usepackage{epsfig}
\usepackage[table,xcdraw,dvipsnames]{xcolor}
\usepackage{caption}
\usepackage{float}
\usepackage{placeins}
\usepackage{color, colortbl}
\usepackage{stfloats}
\usepackage{enumitem}
\usepackage{tabularx}
\usepackage{xstring}
\usepackage{multirow}
\usepackage{xspace}
\usepackage{url}
\usepackage{subcaption}
\usepackage{xcolor}
\usepackage[hang,flushmargin]{footmisc}
\usepackage{setspace}

\ifcamera \usepackage[accsupp]{axessibility} \fi

\newcommand{\nbf}[1]{{\noindent \textbf{#1.}}}

\ifarxiv  \fi

\newcommand{\R}[1]{{%
    \textbf{%
        \ifstrequal{#1}{1}{\textcolor{red}{R#1}}{%
        \ifstrequal{#1}{2}{\textcolor{blue}{R#1}}{%
        \ifstrequal{#1}{3}{\textcolor{magenta}{R#1}}{%
        \ifstrequal{#1}{4}{\textcolor{teal}{R#1}}{%
                           \textcolor{cyan}{R#1}%
        }}}}%
    }%
}}

\DeclareMathOperator*{\argmin}{arg\,min}

\usepackage{xr-hyper}

\makeatletter
\newcommand*{\addFileDependency}[1]{
  \typeout{(#1)}
  \@addtofilelist{#1}
  \IfFileExists{#1}{}{\typeout{No file #1.}}
}

\makeatother

\definecolor{cvprblue}{rgb}{0.21,0.49,0.74}
\usepackage[pagebackref,breaklinks,colorlinks,citecolor=cvprblue]{hyperref}
\usepackage[capitalize]{cleveref}
\crefname{section}{Sec.}{Secs.}
\crefname{table}{Table}{Tables}
\crefname{figure}{Fig.}{Figs.}

\begin{document}
\title{\paperTitle}
\author{\authorBlock}
\maketitle

\begin{abstract}
Learning generalizable visual representations from Internet data has yielded promising results for robotics.
Yet, prevailing approaches focus on pre-training 2D representations, being sub-optimal to deal with occlusions and accurately localize objects in complex 3D scenes. Meanwhile, 3D representation learning has been limited to single-object understanding.
To address these limitations, we introduce a novel 3D pre-training framework for robotics named \textbf{SUGAR} that captures semantic, geometric and affordance properties of objects through 3D point clouds.
We underscore the importance of cluttered scenes in 3D representation learning, and automatically construct a multi-object dataset benefiting from cost-free supervision in simulation.
SUGAR employs a versatile transformer-based model to jointly address five pre-training tasks, namely cross-modal knowledge distillation for semantic learning, masked point modeling to understand geometry structures, grasping pose synthesis for object affordance, 3D instance segmentation and referring expression grounding to analyze cluttered scenes.
We evaluate our learned representation on three robotic-related tasks, namely, zero-shot 3D object recognition, referring expression grounding, and language-driven robotic manipulation.
Experimental results show that SUGAR's 3D representation outperforms state-of-the-art 2D and 3D representations.
\end{abstract}

\vspace{-1em}

\section{Introduction}
\label{sec:intro}

Visual perception plays an essential role in robotics and enables autonomous agents to understand and interact with their physical environment.
Nevertheless, learning generalizable visual representations for robotics is challenging due to the scarcity of real robot data and the large variety of real-world scenes.
Despite substantial efforts in robot data accumulation~\cite{walke2023bridgedata,bharadhwaj2023roboagent,vuong2023openxe} and augmentation~\cite{kostrikov2020imageaug,yu2023robotimginedexp}, it remains prohibitively expensive to collect large-scale datasets comprising a broad range of robotic tasks.

To alleviate the burden of data collection, recent endeavors~\cite{radosavovic2023mvp,nair2023r3m,majumdar2023vc1,khandelwal2022clip4embodied,ma2022vip,karamcheti2023voltron} have sought to leverage large-scale internet data to pre-train 2D visual representations for robotics.
For example, MVP~\cite{radosavovic2023mvp}, VIP~\cite{ma2022vip} and VC-1~\cite{majumdar2023vc1} use self-supervised learning on image or video datasets, while EmbCLIP~\cite{khandelwal2022clip4embodied}, R3M~\cite{nair2023r3m} and Voltron~\cite{karamcheti2023voltron} further perform cross-modal pre-training based on videos with aligned language descriptions.
While these 2D representations have demonstrated promising performance, they still fall short in addressing occlusions in complex cluttered scenes~\cite{wang2021ocidref} and accurately predicting robotic actions~\cite{chen2023polarnet} in the 3D world.

\begin{figure}
    \centering
    \includegraphics[width=\linewidth]{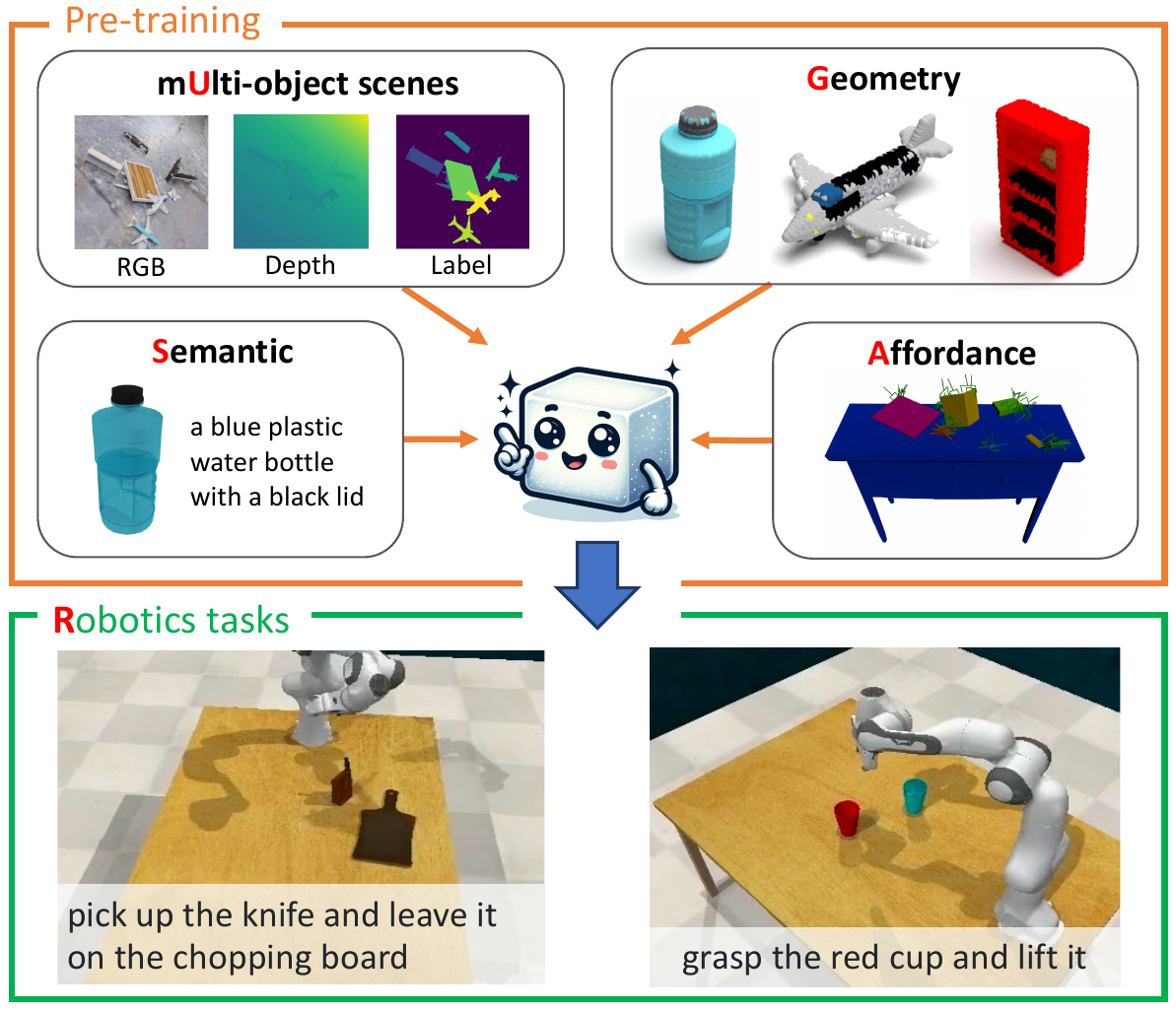}
    \vspace{-.5cm}
    \caption{We introduce SUGAR \includegraphics[height=8pt]{figs/sugar_logo}, a pre-training framework  for robotic-related tasks, which learns semantic, geometry and affordance on both single- and multi-object scenes.
    \vspace{-.5cm}
    }
    \label{fig:teaser}
\end{figure}

Recently, growing research attention has been paid to 3D visual representations for robotics.
A majority of approaches train 3D-based models from scratch~\cite{liu2022framemining,shridhar2023peract,chen2023polarnet}, potentially losing the generalization ability.
Several recent works lift pre-trained 2D features to the 3D space~\cite{jatavallabhula2023conceptfusion,gervet2023act3d,peng2023openscene,zhang2023sgfusion}, which compromise efficiency due to processing multi-view images and do not fully take advantage of 3D data.
To improve 3D representation learning, prior endeavours propose self-supervised pre-training of 3D models~\cite{yu2022pointbert,pang2022pointmae,qi2023recon,liu2023openshape}. 
Pre-training in existing work, however, is typically limited to single objects and complete point clouds, hence, ignoring self-occlusions and clutter in 3D images of real scenes.
In addition, these 3D representations mainly focus on learning geometry and semantic meaning of objects and ignore object affordances which is important for robotics manipulation.

To enhance the capability of 3D representation in robotics, we propose SUGAR - a novel pre-training framework that learns semantics, geometry and affordance properties of objects in 3D point clouds, as illustrated in Figure~\ref{fig:teaser}.
We automatically construct cluttered scenes with multiple objects using large-scale 3D datasets for pre-training, where the object semantics, locations, and grasping poses can be obtained for free in simulation.
To jointly train multiple properties, we propose a versatile transformer-based model comprising a point cloud encoder and a prompt-based decoder.
We use masked point modeling and cross-modal knowledge distillation tasks to train representations for geometry and semantic understanding respectively.
In order to better understand objects and their spatial relations in cluttered scenes, we apply 3D instance segmentation and referring expression grounding tasks in pre-training.
Furthermore, a grasping pose synthesis task is proposed to enable the learning of object affordance in cluttered scenes, which is high-relevant to robotic manipulation. 
We adopt curriculum learning to progressively train SUGAR on single- and multi-object scenes.

We address three downstream tasks for a comprehensive robotic-related evaluation of SUGAR.
The first task is zero-shot 3D object recognition~\cite{liu2023openshape}, a benchmark task for 3D shape understanding;
the second task is referring expression grounding~\cite{wang2021ocidref,lu2023roborefit} in cluttered scenes, which serves as a precursor for interactive robot grasping~\cite{lu2023roborefit}; 
the last but the most important evaluation is language-guided robotic manipulation~\cite{james2020rlbench} that aims to learn a unified policy for multiple robotic tasks (see bottom of Figure~\ref{fig:teaser}).
Experimental results show that SUGAR significantly outperforms models trained from scratch and previous pre-trained models~\cite{nair2023r3m,radford2021clip,karamcheti2023voltron}, demonstrating the importance of 3D pre-training in cluttered scenes and learning object affordances for robotics. 

\noindent In summary, the contributions of our work are three-fold:
\parskip=0.1em
\begin{itemize}[itemsep=0.1em,parsep=0em,topsep=0em,partopsep=0em]
    \item We present SUGAR - a framework with versatile transformer architecture for 3D point cloud representation learning on cluttered scenes.
    \item We pre-train SUGAR on five tasks, namely masked point modeling, cross-modal learning, grasping pose synthesis, instance segmentation and object grounding, enabling to learn semantics, geometry, and affordance of objects. 
    \item We experimentally demonstrate that SUGAR outperforms the state of the art on three robotic-related tasks including zero-shot object recognition, referring expression grounding and language-guided robotic manipulation.
\end{itemize}

\vspace{-0.5em}
\section{Related Work}
\label{sec:related}
\vspace{-0.5em}
\noindent\textbf{Visual representation learning for robotics} is a fundamental yet challenging problem.
To learn the representation, many approaches~\cite{brohan2022rt1,jang2022bcz,bharadhwaj2023roboagent,radosavovic2023sensorimotorpretrain} rely on in-domain data from target robotic tasks such as real robot demonstrations.
While significant efforts have been made to collect more real robot data~\cite{walke2023bridgedata,bharadhwaj2023roboagent,vuong2023openxe}, the scalability of such data is still constrained by its cost.
To alleviate the data scarcity issue, various techniques have been explored including data augmentation~\cite{laskin2020rlaugdata,kostrikov2020imageaug,yu2023robotimginedexp,bharadhwaj2023roboagent}, self-supervised learning~\cite{laskin2020curl,pari2021repforil,radosavovic2023sensorimotorpretrain},
and the integration of task-specific information~\cite{jonschkowski2015learningprior,zhang2020learninginvariantrep}.
More recently, thanks to the advancement in pre-training with large-scale internet data~\cite{sun2019videobert,he2022mae,radford2021clip}, a number of works~\cite{shah2021rrl,khandelwal2022clip4embodied,parisi2022pretrainvision4robo,nair2023r3m,radosavovic2023mvp,xiao2022mvp4control,ma2022vip,majumdar2023vc1,dasari2023datasets4robots,karamcheti2023voltron} have showcased the benefits of applying pre-trained visual representations to the robotic domain such as features learned on ImageNet~\cite{deng2009imagenet}, videos of human performing everyday tasks~\cite{goyal2017something,grauman2022ego4d}, and CLIP~\cite{radford2021clip} features.
In particular, Voltron~\cite{karamcheti2023voltron} demonstrates the advantage of incorporating aligned language descriptions to the visual representation learning for language-driven robotic tasks.
Nevertheless, these approaches predominantly pre-train 2D visual representations, which can be sub-optimal for robotics in the 3D world.
In this work, we focus on pre-training a 3D representation for robotic-related tasks.

\noindent\textbf{3D point cloud understanding} has received increased attentions with the rapid development of 3D sensors~\cite{guo2020pcdsurvey}.
A variety of neural network architectures have been developed to effectively process 3D point cloud data such as voxel-based networks~\cite{riegler2017octnet}, Graph CNNs~\cite{wang2019dgcnn} and PointNet series~\cite{qi2017pointnet,qi2017pointnet++,qian2022pointnext}. 
As self-attention in transformers~\cite{vaswani2017attention} is fundamentally a set operator, transformer-based models~\cite{guo2021pct,zhao2021pointtransformer} have gained popularity for unordered points and achieved superior performance in 3D object recognition~\cite{liu2023openshape}, scene segmentation~\cite{schult2023mask3d} and so on.
The transformer architecture also makes it easy to pre-train by self-supervised masked point modeling~\cite{yu2022pointbert,pang2022pointmae} or cross-modal learning from images and texts~\cite{xue2023ulip,dong2022autoencoders,zhang2023i2pmae,xue2023ulip2,liu2023openshape}.
ReCon~\cite{qi2023recon} further proposes a new transformer architecture to benefit from multiple pre-training tasks for point cloud understanding.
However, these works only pre-train 3D representation on complete point clouds of single objects, limiting the generalization ability to more complex scenes.
Ponder~\cite{huang2023ponder} is a pioneer work to pre-train point clouds of indoor scenes using neural rendering, but it ignores the semantic information in representation learning.
Furthermore, none of the above works considers affordance learning for 3D objects which has particular importance for robotic manipulation.
Our proposed approach SUGAR is designed to address these limitations by jointly pre-training on five diverse tasks in multi-object settings.

\noindent\textbf{Robotic manipulation} refers to the control of environment through selective contacts by the agent~\cite{mason2018toward}.
This task requires rich perceptual and common-sense understanding of objects around the robot.
Our work concentrates on two important manipulation problems: object grasping - a fundamental skill for robotic manipulation~\cite{kleeberger2020surveygrasp,newbury2023deepgraspsurvey}, and language-guided robotic manipulation~\cite{stepputtis2020languageroboman,ichter22saycan,lynch22interactlang,zhu22viola} which enables convenient human-robot interaction and skill generalization across tasks.
Many deep learning based grasping methods~\cite{li2021simultaneous,sundermeyer2021contact} are based on 3D point clouds due to its superiority to address visual occlusions and good sim-to-real transfer performance. 
Training such models has been facilitated by large-scale grasping datasets, for example, GraspNet-1Billion~\cite{fang2020graspnet} contains real RGB-D scans and simulated grasp poses and  ACRONYM~\cite{eppner2021acronym} is composed of synthetic scenes and grasps verified in physical simulation.
For language-guided robotic manipulation, though 3D visual representations have recently been explored~\cite{james22c2f-arm,shridhar2023peract,chen2023polarnet}, most existing methods still rely on 2D visual representations~\cite{driess2023palme,brohan2022rt1,shridhar21cliport,reed2022gato,gervet2023act3d} in order to benefit from pre-trained vision-and-language models.
To the best of our knowledge, our work is the first to explore large-scale 3D pre-training techniques to improve robotic manipulation.

\section{SUGAR: 3D Pre-training for Robotics}
\label{sec:method}

We propose SUGAR, a pre-training framework for 3D point clouds which learns semantic, geometric and affordance properties of objects during pre-training.
We first introduce the network architecture in Section~\ref{sec:method_network} and then describe the self-supervised pre-training tasks in Section~\ref{sec:method_pretrain_tasks}.
In Section~\ref{sec:method_pretrain_data}, we present the construction of pre-training datasets, followed by implementation details in  Section~\ref{sec:method_pretrain_details}.

\subsection{Network Architecture}
\label{sec:method_network}

\begin{figure}
    \centering
    \includegraphics[width=\linewidth]{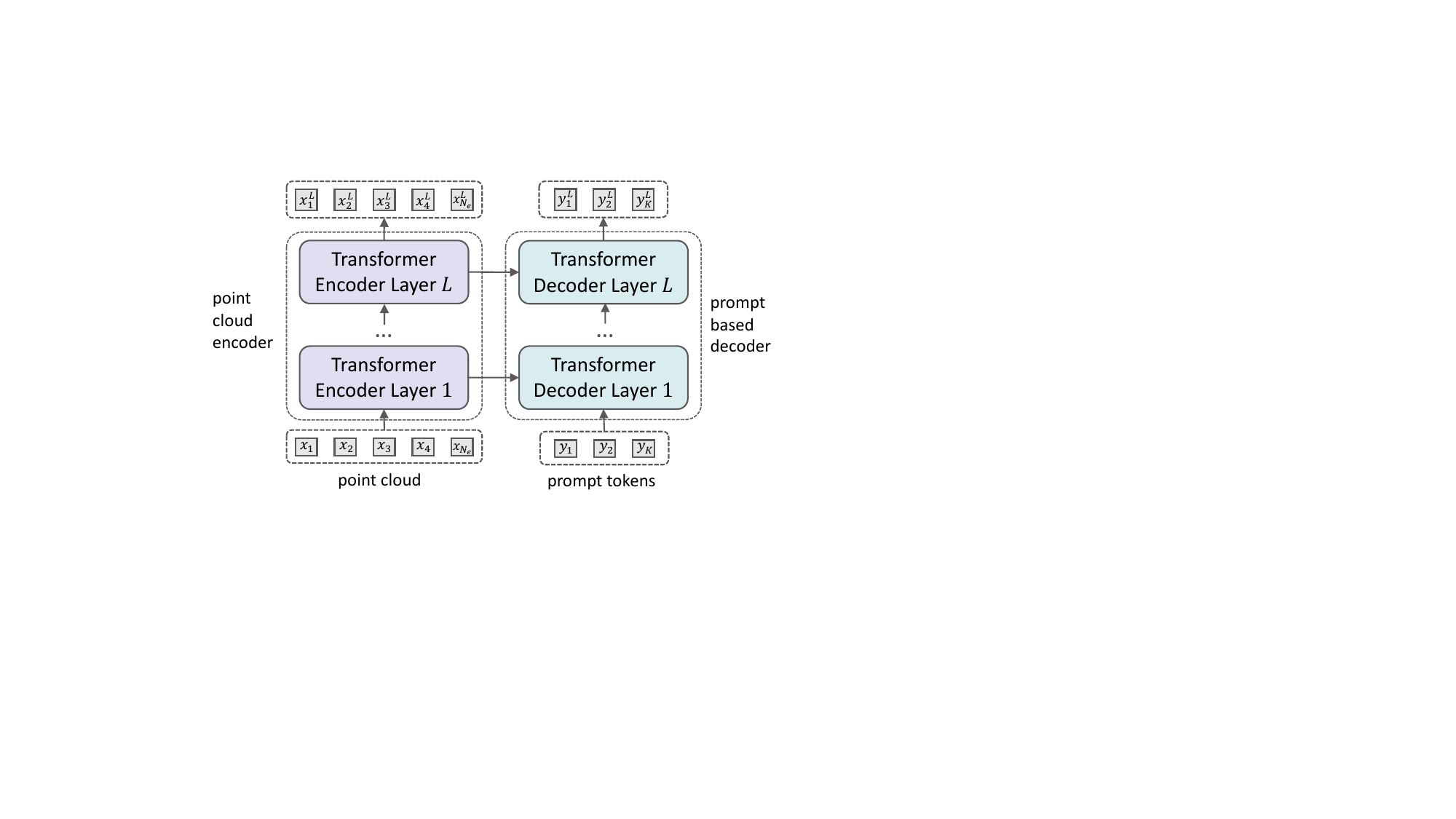}
    \caption{Network architecture of SUGAR. It consists of a point cloud encoder to generate point embeddings and a prompt-based decoder that takes task-specific prompt tokens and layer-wise connections to point embeddings to obtain prompt embeddings.}
    \label{fig:network_arch}
    \vspace{-1em}
\end{figure}

Figure~\ref{fig:network_arch} illustrates our network architecture, which is a versatile framework to solve multiple tasks given the input point cloud and task-specific prompts.
Suppose $X=\{x_i\}_{i=1}^{N}$ is a point cloud where $x_i \in \mathbb{R}^{6}$ is composed of 3D coordinates and RGB colors and $N$ is the number of points, and $Y=\{y_i\}_{i=1}^{K}$ is a sequence of prompt tokens (explained in Section~\ref{sec:method_pretrain_tasks}).
The model consists of a point cloud encoder $\mathcal{E}(X)$ to generate point embeddings and a prompt-based decoder $\mathcal{D}(\mathcal{E}(X), Y)$ to obtain task-specific embeddings, both of which utilizes transformer blocks~\cite{vaswani2017attention}. We present the details of the two modules below.

\nbf{Point cloud encoder}
Given $X$, we first use farthest point sampling to select $N_e$ key points and group $S_e$ nearest points for each key point as a local point cloud.
We normalize the local point cloud using the corresponding key point as the center and employ a shared PointNet~\cite{qi2017pointnet} to encode each local point cloud into a token embedding $x^0_i \in \mathbb{R}^{d}$ where $d$ is the dimensionality of the feature.
The position of the local point cloud is set as the 3D coordinates of its key point, and a feed-forward network (FFN) is utilized to obtain the position embedding $x^p_i \in \mathbb{R}^{d}$.
We use a standard transformer model with $L$ layers to encode the point cloud tokens. The computation at the $l$-th layer is:
\begin{equation}
    \{x^{l}_i\}_{i=1}^{N_e} = \text{FFN}(\text{SA}(\{x^{l-1}_i + x^p_i\}_{i=1}^{N_e})),
\end{equation}
where $\text{SA}$ is the self-attention operator. We omit the residual and layer normalization for simplicity. Please refer to the transformer paper~\cite{vaswani2017attention} for details.

\nbf{Prompt-based decoder}
Given the task-specific prompt tokens $Y$, we use a linear layer to project them into token embeddings $\{y^0_i\}_{i=1}^{K}, y^0_i \in \mathbb{R}^d$ where the linear layer is different for each task.
The decoder consists of the same number of blocks of self-attention (SA) and cross-attention (CA) layers as the encoder to layer-wisely query the encoded point embeddings and update the prompt tokens:
\begin{align}
    \{\hat{y}^{l}_i\}_{i=1}^{K} &= \text{FFN}(\text{SA}(\{y^{l-1}_i\}_{i=1}^{K})), \\
    \{y^{l}_i\}_{i=1}^{K} &= \text{FFN}(\text{CA}(\{\hat{y}^{l}_i\}_{i=1}^{K}, \{x^{l}_i\}_{i=1}^{N_e})).
\end{align}
The output embeddings $\{y^L_i\}_{i=1}^{K}$ can then be used for each specific task described below.

\subsection{Pre-training Tasks}
\label{sec:method_pretrain_tasks}

\begin{figure*}
    \centering
    \includegraphics[width=\linewidth]{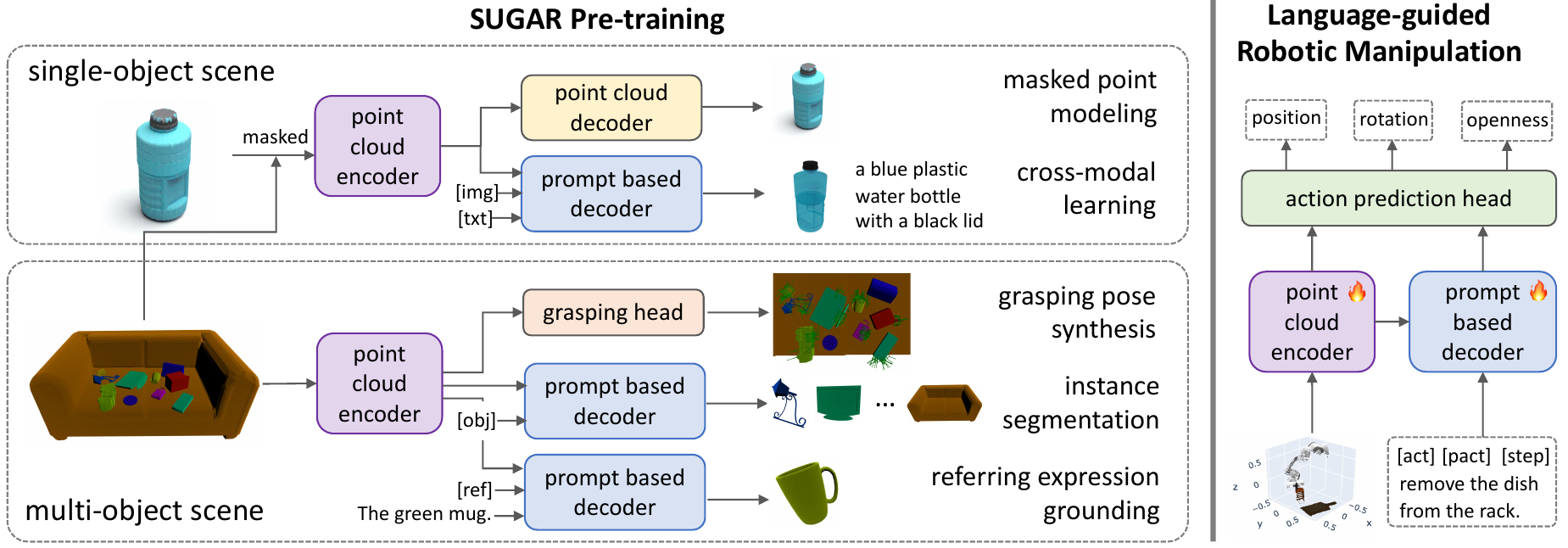}
    \caption{Left: Five pre-training tasks for SUGAR using single- and multi-object scenes. The modules of the same color are shared. Right:  The pre-trained point cloud encoder and prompt-based decoder are finetuned 
    on the downstream task of robotic manipulation.
    }
    \label{fig:pretrain_tasks}
    \vspace{-1em}
\end{figure*}

In this section, we describe the five pre-training tasks that empowers SUGAR to recognize, segment and grasp objects in 3D point clouds and, hence, to be a state-of-the-art 3D representation for robotic manipulation. 
Figure~\ref{fig:pretrain_tasks} (left) illustrates all the pre-training tasks.

\nbf{Masked point modeling (MPM)}
Masked modeling is a general self-supervised task in many different domains such as texts~\cite{devlin2018bert}, images~\cite{he2022mae}, videos~\cite{sun2019videobert} and 3D point clouds~\cite{yu2022pointbert,pang2022pointmae,qi2023recon}.
It enables learning geometry structures by masking a fraction of points and reconstructing the original point cloud.
Specifically, we randomly mask out 60\% of local point cloud tokens following the best practice in~\cite{qi2023recon} and only feed the unmasked tokens to the point cloud encoder $\mathcal{E}$.
We then reconstruct the masked tokens via a light-weight point cloud decoder which is a 4-layer plain transformer. 
The decoder is fed with unmasked point embeddings $\{x^L_i\}$ and a special \verb|[mask]| embedding added to the corresponding position embedding for each masked token.
The output embeddings of the masked tokens are utilized to predict 3D coordinates of each point in the local point cloud. We further include color prediction for each point to enhance the learning of textures.
Assume $\hat{X}_j \in \mathbb{R}^{S_e \times 6}$ and $X_j \in \mathbb{R}^{S_e \times 6}$ are the predicted and groundtruth local point cloud respectively, we extend the original reconstruction loss $l_2$ Chamfer Distance~\cite{fan2017chamferdist} to also consider colors:
\begin{equation}
\footnotesize
\begin{aligned}
    &L_{mpm} = \frac{1}{N_e S_e} \sum_{j=1}^{N_e} (\sum_{(\hat{x}, x) \in A_j} ||\hat{x} - x||_2^2 + \sum_{(x, \hat{x}) \in B_j} ||x - \hat{x}||_2^2 ) \\
    &\text{where} \\
    &A_j=\{(\hat{x}, x)| \hat{x} \in \hat{X_j}, x \in X_j, x = \argmin_{x \in X_j} ||\hat{x}_{:3}-x_{:3}||_2^2\}, \\
    &B_j=\{(x, \hat{x})| x \in X_j, \hat{x} \in \hat{X_j}, \hat{x} = \argmin_{\hat{x} \in \hat{X}_j} ||x_{:3}-\hat{x}_{:3}||_2^2\}.
\end{aligned}
\end{equation}
The $x_{:3}$ denotes the first three dimension of the vector $x$, which is the 3D coordinates of the point.

\nbf{Cross-modal learning (CML)}
As pre-trained image and text models have achieved great success in representation learning, it is beneficial to distill knowledge from these models to train the 3D visual representation~\cite{xue2023ulip,qi2023recon,xue2023ulip2,liu2023openshape}. %
Suppose we have $R^I$ images and $R^T$ text descriptions aligned with a point cloud $X$, for instance, by projecting $X$ from 3D to 2D using different camera poses and captioning the 2D images.
We use pre-trained image and text models to extract image and text features as $\{f^I_r\}_{r=1}^{R^I}$ and $\{f^T_r\}_{r=1}^{R^T}$ respectively.
Our goal is to extract point cloud features from $X$ that are close to the aligned image or text features. For this purpose, we leverage two prompt tokens \verb|[img]| and \verb|[txt]| as input to the decoder $\mathcal{D}$ and then project the two output embeddings via a linear layer to the same space as the pre-trained image or text features, denoted as $\hat{f}^I$ and $\hat{f}^T$.
A smoothed $l_1$ loss $L_{cml}$ is used to minimize the distance between $\hat{f}^I$ and each $f^I_r$, as well as between $\hat{f}^T$ and each $f^T_r$ for cross-modal knowledge distillation.

\nbf{Grasping pose synthesis (GPS)}
Object grasping is a precursor for many robotic applications and thus we treat it as a fundamental skill to learn for the visual representation. 
It is however challenging to predict diverse grasping poses through a deterministic model.
To simply the problem, we make an assumption that there exists at most one optimal grasping pose for each point in $X$, where $g^v_i \in \{0,1\}$ denotes if the point has a valid grasping pose and $g^m_i \in \mathbb{R}^{4 \times 4}$ is the optimal grasping pose for the point if $g^v_i = 1$.
Given point embeddings $\{x^L_i\}_{i=1}^{N_e}$ from $\mathcal{E}$, we first use tricubic interpolation to upsample features for each point $\{x^L_i\}_{i=1}^{N}$.
Then we use a linear layer to get the binary prediction $\hat{g}^v_i \in \mathbb{R}$ given $x^L_i$.
For valid points, we predict the relative position of the optimal grasping pose to the point as $\hat{g}^{p}_i \in \mathbb{R}^3$ and the 6D representation $\hat{g}^{r}_i \in \mathbb{R}^6$ for 3D rotation~\cite{zhou2019continuity}, which can form grasping pose $\hat{g}^m_i \in \mathbb{R}^{4 \times 4}$ using Gram-Schmidt process.
The training objective consists of a binary cross-entropy loss (BCE) and a $l_2$ distance for the grasping pose:
\begin{equation}
\small
    L_{gps} = \frac{1}{N} \sum_{i=1}^{N} \text{BCE}(\hat{g}^v_i, g^v_i) + g^v_i ||\hat{g}^m_i - g^m_i||_2^2.
\end{equation}

\nbf{Instance segmentation (INS)}
Segmenting 3D objects is a key ability to understand cluttered scenes.
To address the task, we use a set of object tokens (\verb|[obj1]|, $\cdots$, \verb|[objK]|) as the prompt tokens to the decoder $\mathcal{D}_\theta$ and obtain output embeddings $\{y^L_i\}_{i=1}^{K}$. 
For each object token $y_i$, we measure its similarity with the upsampled point embeddings $s(y_i, x_j) = y^L_i \cdot x^L_j$ and thus generate the instance segmentation mask $m_{i} \in \mathbb{R}^{N}, m_{ij} = \sigma(s(y_i, x_j)) > 0.5$ where $\sigma$ is the sigmoid function.
In addition, we use linear layers to predict an objectiveness score given $y^L_i$ and image and text features similar to that in CML task.
To train the model, we take the approach in DETR~\cite{carion2020detr} to use Hungarian matching for groundtruth assignment, and then given the best matches we compute a combined loss including a BCE loss and DICE loss~\cite{sudre2017dice} for instance segmentation, a BCE loss for objectiveness score prediction and a smoothed $l_1$ loss for cross-modal learning on the object level.

\nbf{Referring expression grounding (REG)}
The task aims to segment an object described by a natural language sentence in cluttered scenes.
We use a prompt token \verb|[ref]| and a sequence of encoded text tokens by a pre-trained language model to $\mathcal{D}$, and predict the object mask using the output embedding of \verb|[ref]| similar to the INS task.
We combine the BCE loss and DICE loss for training.

\subsection{Pre-training Data}
\label{sec:method_pretrain_data}

We first introduce the single-object datasets and then describe our automatic constructed multi-object dataset.

\nbf{Single-object dataset}
ShapeNet~\cite{chang2015shapenet} is a commonly used dataset for 3D point cloud pre-training~\cite{yu2022pointbert,pang2022pointmae,qi2023recon}, containing 55 categories and about 52K instances.
To scale up the pre-training dataset, we follow OpenShape~\cite{liu2023openshape} to ensemble ShapeNet, 3D-FUTURE~\cite{fu20213dfuture} (16.5K instances), ABO~\cite{collins2022abo} (8.0K instances) and Objaverse~\cite{deitke2023objaverse} (no LVIS split, 752.2K instances).
We use the same point clouds, aligned images and text descriptions as~\cite{liu2023openshape}.
Specifically, the point cloud of an object is obtained by evenly sampling 10K points from the mesh surface of 3D object asset and interpolating the point colors based on the mesh texture.
The aligned images are generated by rendering the object from 12 fixed views around the object, and the text descriptions are built by captioning the rendered images via existing image captioning systems, retrieving descriptions of similar images, or the metadata of the object name. 

\nbf{Multi-object dataset}
We use Blenderproc~\cite{Denninger2023blenderproc} for  automatic multi-object scene construction. 
First, we randomly select 2 to 10 objects from  single-object datasets and sample their locations in a 3D space. 
Then, we enable the physical simulation to fall objects onto a plane with random textures to generate a realistic scene. 
Finally, we randomly sample camera position near the scene and let the camera point to the center of the scene for RGB-D image and segmentation label rendering.
Figure~\ref{fig:teaser} top left shows an example of the generated data.
The point cloud and per-point labels can be obtained by projecting the 2D image to 3D given the camera parameters.
In total, we generate 48.9K multi-object scenes using objects from ShapeNet, and 62.8K scenes using objects from Objaverse (no LVIS) dataset.

\noindent\emph{Grasping pose generation.} It is expensive to automatically generate high-quality grasping poses requiring pose sampling and verification in physical simulation~\cite{eppner2021acronym}. 
However, recent works~\cite{fang2020graspnet,wan2023unidexgrasp++} have shown that having diverse grasp poses per objects is more important than scale to generalize to various objects. 
Therefore, we re-use the existing grasping dataset ACRONYM~\cite{eppner2021acronym} which contains 2K physically verified grasping poses per object for around 8K objects in ShapeNet. 
We generate 62.7K multi-object scenes following~\cite{eppner2021acronym} and filter invalid grasp poses with collisions in the scene.
For each point, we select the nearest grasping pose to the point as the optimal grasping pose; there is no optimal pose if the nearest distance is above a certain threshold.

\subsection{Pre-training Details}
\label{sec:method_pretrain_details}

\noindent\textbf{Training strategy.}
We adopt a curriculum learning scheme to pre-train the 3D representations as it is more challenging to understand the multi-object scenes than single objects.
We first pre-train the model on the single-object datasets with the MPM and CML tasks as show in the top row of Figure~\ref{fig:pretrain_tasks}~(left); then we joint train on both single- and multi-object datasets using all the five pre-training tasks.

\noindent\textbf{Implementation details.}
We set the number of points $N=4096$, the number of key points $N_e=256$ and the group size $S_e=32$.
Due to the computational cost, we only adopt a small model size with $d=384$, $L=12$.
In the CML task, OpenCLIP ViT-bigG-14~\cite{openclip} is used to extract image and text features.
For the multi-object dataset, the plane background is kept in the point cloud 50\% of the time during training.
We pre-train two sets of models according to the pre-training data: `SN' uses objects only in ShapeNet, and `Ens' uses the ensembled four datasets.
More training details are presented in the supplementary material.

\section{Evaluation on Robotic-related Tasks}
\label{sec:expr}

To thoroughly evaluate the pre-trained representation, we resort to three robotic-related tasks including zero-shot object recognition, referring expression grounding, and language-guided robotic manipulation. We present datasets, downstream adaptation and quantitative results for each task in the following three sections.

\subsection{Zero-shot Object Recognition}

\begin{table*}
\centering
\small
\caption{Zero-shot object recognition performance on three benchmarks. The Top1 accuracy is reported if not specified otherwise. The blue colored results in brackets on the ScanObjectNN dataset are obtained using colored point clouds. 
\vspace{-.2cm}
}
\label{tab:zeroshot_cls}
\begin{tabular}{llccccccc} \toprule
\multirow{2}{*}{Pretrain data} & \multirow{2}{*}{Method} & \multirow{2}{*}{ModelNet40} & \multicolumn{3}{c}{ScanObjectNN} & \multicolumn{3}{c}{Objaverse-LVIS} \\ 
 &  &  & OBJ\_ONLY & OBJ\_BG & PB\_T50\_RS & Top1 & Top3 & Top5 \\ \midrule
\multirow{6}{*}{ShapeNet} & ReCon~\cite{qi2023recon} & 61.2 & 39.6 & 38.0 & 29.5 & 1.1 & 2.7 & 3.7 \\
 & CLIP2Point~\cite{huang2023clip2point} & 49.5 & 35.5 & 30.5 & 23.3 & 2.7 & 5.8 & 7.9 \\
 & ULIP-PointBERT~\cite{xue2023ulip} & 60.4 & 49.9 & - & - & 6.2 & 13.6 & 17.9 \\
 & OpenShape-PointBERT~\cite{liu2023openshape}\footnotemark[1]  & 70.3 & 51.8 (\textcolor{blue}{52.0}) & 41.9 (\textcolor{blue}{42.4}) & 28.5 (\textcolor{blue}{28.6}) & 10.8 & 20.2 & 25.0 \\ \cmidrule{2-9}
\rowcolor{gray!15} \cellcolor{white} & SUGAR (single) & 71.2 & 48.6 (\textcolor{blue}{52.8}) & 46.3 (\textcolor{blue}{50.3}) & 33.4 (\textcolor{blue}{35.4}) & 13.3 & 22.6 & 27.3 \\
\rowcolor{gray!15} \cellcolor{white} & SUGAR (multi) & 66.5 &  \textbf{53.5} (\textcolor{blue}{55.0}) &  47.9 (\textcolor{blue}{50.8}) &  34.2 (\textcolor{blue}{36.5}) & 12.1 & 20.1 & 25.1 \\ \midrule
\multirow{4}{*}{\begin{tabular}[c]{@{}l@{}}Ensembled \\ (no LVIS)\end{tabular}} & ULIP-PointBERT~\cite{xue2023ulip} & 71.4 & 46.0 & - & - & 21.4 & 38.1 & 46.0 \\
 & OpenShape-PointBERT~\cite{liu2023openshape}\footnotemark[1] & \textbf{85.3} & 50.8 (\textcolor{blue}{52.0}) & 51.4 (\textcolor{blue}{52.6}) & \underline{39.4} (\textcolor{blue}{40.3}) & 39.1 & 60.8 & 68.9 \\ \cmidrule{2-9}
 &  \cellcolor{gray!15}SUGAR (single) &   \cellcolor{gray!15}\underline{84.3} &   \cellcolor{gray!15}49.6 (\textcolor{blue}{\textbf{65.3}}) &  \cellcolor{gray!15} \textbf{56.2} (\textcolor{blue}{\textbf{68.0}}) &  \cellcolor{gray!15} \textbf{41.8}
 (\textcolor{blue}{\textbf{49.3}}) &  \cellcolor{gray!15}\textbf{42.1} &  \cellcolor{gray!15}\textbf{64.6} &  \cellcolor{gray!15}\textbf{72.1} \\
\rowcolor{gray!15} \cellcolor{white} & SUGAR (multi) & 83.8 & \underline{53.2} (\textcolor{blue}{\underline{63.5}}) & \underline{53.2}  (\textcolor{blue}{\underline{65.6}}) & 38.2 (\textcolor{blue}{\underline{48.7}})& \underline{39.4} & \underline{61.6} & \underline{69.3}
 \\ \bottomrule
\end{tabular}
\vspace{-.2cm}
\end{table*}

The task aims to classify unseen 3D objects without training on those specific categories.
It evaluates the generalization ability of visual representations for semantic understanding.

\noindent\textbf{Datasets.}
We use three 3D object recognition benchmarks including ModelNet40~\cite{wu2015modelnet}, ScanObjectNN~\cite{uy2019scanobjectnn} and Objaverse-LVIS~\cite{deitke2023objaverse}.
ModelNet40 contains 40 categories and 2,468 objects in the test split. The objects are synthetic 3D models without colors.
ScanObjectNN is one of the most challenging 3D datasets, consisting of 15 common categories and 587 real-world 3D scans in the test split. There are three evaluation setups with increasing levels of difficulty: 1)~OBJ\_ONLY which only includes ground truth segmented objects; 2)~OBJ\_BG where objects are with background data; 3)~PB\_T50\_RS with 2,935 testing examples where perturbations are applied to the ground truth object bounding boxes to extract the point cloud. The scanned objects contain colors.
Objaverse-LVIS~\cite{deitke2023objaverse} is an annotated subset of the large-scale Objaverse dataset and comprises 46,205 shapes among 1,156 LVIS~\cite{gupta2019lvis} categories. The objects are crawled from Internet, containing both synthetic 3D models and real-world 3D scans. Colors are included in the point cloud.

\noindent\textbf{Evaluation metrics.}
We use Top1 classification accuracy as the main evaluation metric, and also report Top3 and Top5 for the Objaverse dataset to compare with previous work.

\footnotetext[1]{We re-run OpenShape on the ScanObjectNN as the reported number in OpenShape~\cite{liu2023openshape} is on the whole dataset of the OBJ\_ONLY setup.}

\noindent\textbf{Downstream adaptation.}
We randomly sample 4,096 points for each object and set RGB values as -0.2 (gray color) if there is no color in the point cloud.
As in the CML pre-training task, we use \verb|[img]| and \verb|[txt]| prompt tokens to extract point cloud features that are in the same space of the pre-trained image and text features.
The two predicted features are fused to obtain the final representation, which is used to perform KNN classification with the pre-trained text features of object categories.

\noindent\textbf{Results.}
Table~\ref{tab:zeroshot_cls} presents the performance on the test split of the three benchmarks.
In the top block, all the compared methods and SUGAR (single) are trained on single objects in ShapeNet, while SUGAR (multi) is further trained on multi-object scenes constructed from ShapeNet objects.
Our SUGAR (single) achieves comparable performance with the state-of-the-art method OpenShape~\cite{liu2023openshape} on ModelNet40 and ScanObjectNN without colors, and significantly outperforms OpenShape on Objaverse-LVIS and ScanObjectNN with colors, \eg, +7.9\% on OBJ\_BG split.
As OpenShape randomly discards all point cloud colors during pre-training, it sacrifices the ability of texture recognition but improves the results for uncolored point clouds.
Our model however learns to reconstruct both geometry structures and colors in masked point modeling, 
taking advantage of both geometric and texture information for semantic recognition.
SUGAR (multi) deteriorates the performance on ModelNet40 and Objaverse-LVIS datasets where the point clouds are complete, but performs better on the ScanObjectNN dataset with real scanned objects.
This is because the multi-object dataset captures more realistic scenes with occlusions.
Furthermore, scaling up the pre-training data significantly improves the zero-shot recognition performance as shown in the bottom block of Table~\ref{tab:zeroshot_cls}.
Since the Objaverse dataset contains both synthetic 3D objects and real scans, the performance on ScanObjectNN is boosted by a large margin.
SUGAR (single) achieves 68.0\% Top1 accuracy on OBJ\_BG split of ScanObjectNN, outperforming state of the art by 15.4\%. 
However, we observe around 3\% decrease in SUGAR (multi) compared to SUGAR (single). 
We hypothesize two reasons for the performance drop. First, we only use a small transformer model which may not have sufficient capacity to jointly solve the five pre-training tasks when the pre-training data increases. 
Second, there are multi-object scenes in the crawled Objaverse dataset such as multi-level buildings, hence, construction of new scenes using 3D assets in Objaverse may require additional treatment, which we leave to future work.
For completeness of comparison with prior work~\cite{liu2023openshape}, we include results of pre-training on Ensembled with LVIS dataset in the supplementary material, which show similar trend.

\subsection{Referring Expression Grounding}

Given a natural language description of an object, the task is to segment the target object in the 3D point cloud of cluttered scenes. Solutions to this task require semantic scene understanding and spatial relation reasoning.

\begin{figure*}
    \centering
    \begin{subfigure}[b]{0.45\linewidth}
         \centering
         \includegraphics[width=.95\linewidth]{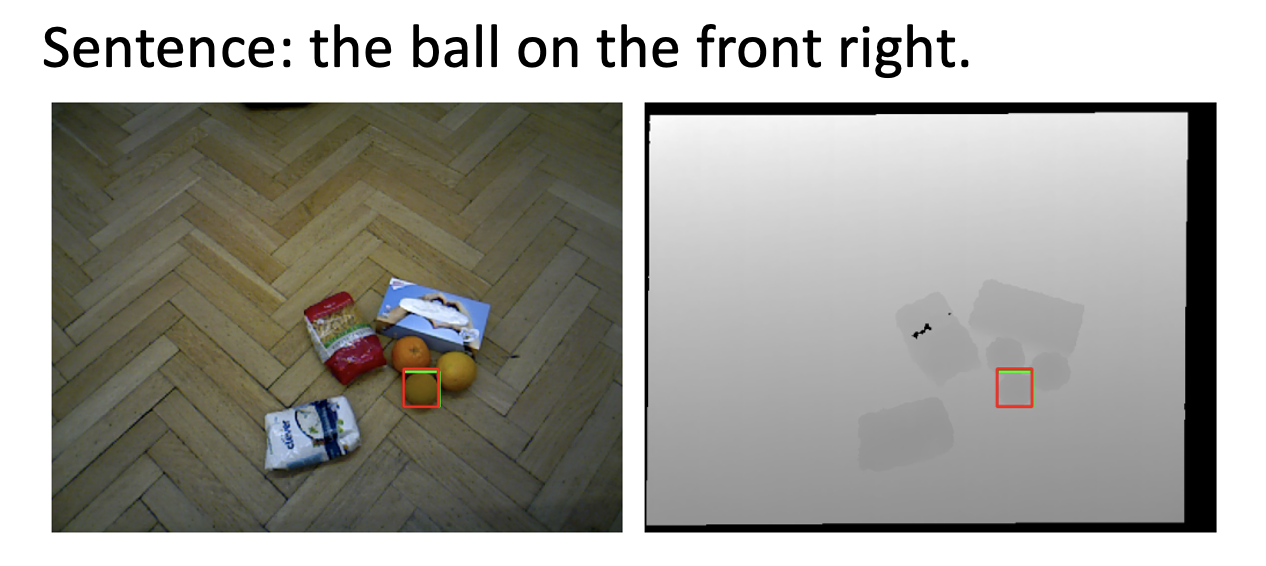}
         \vspace{-.2cm}
         \caption{An example in the test split of OCID-Ref dataset.}
     \end{subfigure}
     \hfill
     \begin{subfigure}[b]{0.45\linewidth}
         \centering
         \includegraphics[width=.95\linewidth]{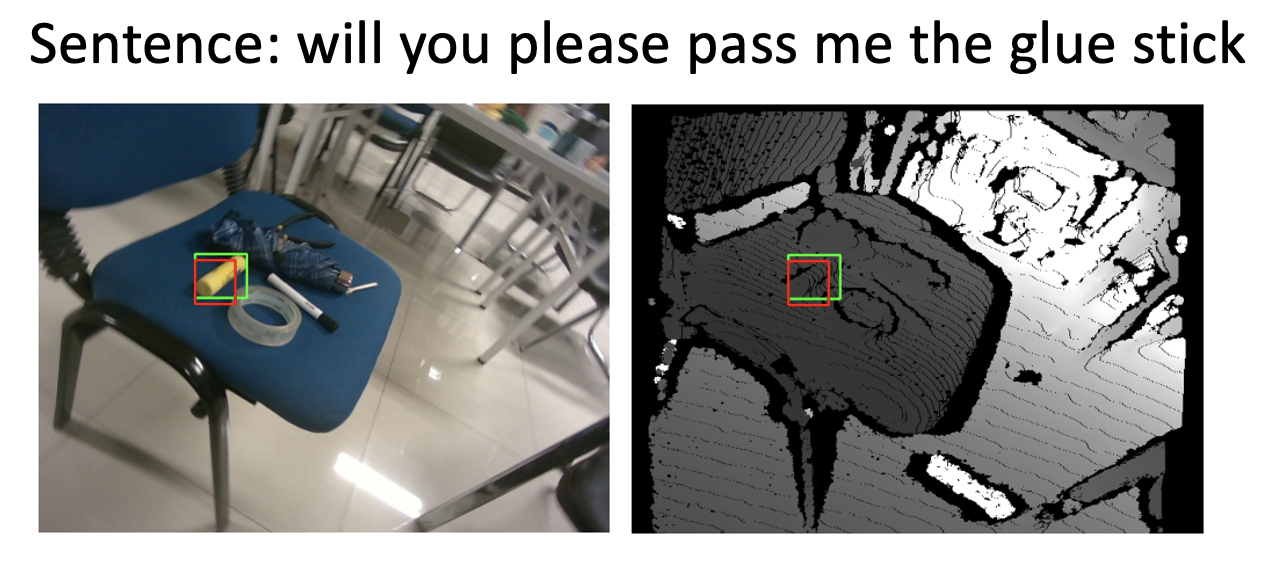}
         \vspace{-.2cm}
         \caption{An example in the testB split of RoboRefit dataset.}
     \end{subfigure}
     \vspace{-.2cm}
    \caption{Referring expression examples on the OCID-Ref and RoboRefit dataset. The green bounding box is the groundtruth annotation, and the red bounding box is predicted by our SUGAR model. RoboRefit contains natural scenes and noisy depth observations.
    \vspace{-.3cm}
    }
    \label{fig:reg_examples}
\end{figure*}

\noindent\textbf{Datasets.}
We use the OCID-Ref~\cite{wang2021ocidref} and  RoboRefit~\cite{lu2023roborefit} datasets, which are representative of robotic manipulation scenes.
OCID-Ref is collected in clean lab environments and consists of 58 object categories, 2,298 RGB-D images and 259,839 referring expressions for training. It has 18,342 and 27,513 referring expressions in validation and test splits respectively. However, all the validation and test images appear in the training split. This makes it impossible to evaluate the generalization ability for unseen scenes and objects.
In contrast, RoboRefit contains more natural and diverse scenes captured by noisy RGB-D cameras. There are 7,929 RGB-D images and 36,915 referring expressions in the training split.
Two test splits are used for evaluation: testA shares similar scenes to the training split with 1,859 scenes and 8,523 sentences; 
scenes and objects of testB with 1,083 RGB-D images and 5,320 sentences are different from training.
Figure~\ref{fig:reg_examples} shows examples of the two datasets.

\noindent\textbf{Evaluation metrics.}
Previous works~\cite{wang2021ocidref,lu2023roborefit,karamcheti2023voltron} all evaluate in the 2D domain. For fair comparison, we first predict a 3D segmentation mask given the point cloud and then project it in 2D.
The major metric in the OCID-Ref dataset is Acc@0.25 which is the ratio of predicted 2D bounding boxes that have IoU larger than 0.25 with the groundtruth. RoboRefit uses Acc@0.5 for 2D object detection and mIoU for segmentation which is the mean IoU between the predicted and groundtruth segmentation masks.

\noindent\textbf{Downstream adaptation.}
We finetune SUGAR similar to the pre-training REG task for 3D segmentation mask prediction, except that we increase the number of key points $N_e$ to improve precision. We set $N_e=512$ for OCID-Ref and $1536$ for RoboRefit if not stated otherwise. 
As the depth sensor in RoboRefit is noisy, we automatically remove outliers~\cite{open3d} to clean 3D object masks in training and evaluation.

\begin{table}
\centering
\small
\tabcolsep=0.3cm
\caption{The results of Acc@0.25 for referring expression detection on the testing split of OCIF-Ref dataset.}
\vspace{-.2cm}
\label{tab:ocidref_results}
\begin{tabular}{lcccc} \toprule
\multirow{2}{*}{Method} & \multicolumn{4}{c}{Clutter level} \\
& Total & Min & Med & Max \\ \midrule
R3M~\cite{nair2023r3m} & 63.30 & 63.87 & 68.34 & 55.33 \\
MVP~\cite{radosavovic2023mvp} & 49.58 & 50.98 & 53.83 & 41.94 \\
CLIP~\cite{radford2021clip} & 68.35 & 67.01 & 76.61 & 60.33 \\
V-Cond~\cite{karamcheti2023voltron} & 90.77 & 87.56 & 96.58 & 90.17 \\ \midrule
SUGAR (Ens\_m) & \textbf{97.74} & \textbf{98.52} & \textbf{97.75} & \textbf{96.68} \\ \bottomrule
\end{tabular}
\vspace{-1.5em}
\end{table}

\begin{table}
\centering
\small
\tabcolsep=0.07cm
\caption{
Performance of referring expression detection (evaluated by Acc@0.5) and referring expression segmentation (evaluated by mIoU) on the RoboRefit dataset.
We use $N_e=1536$ for our SUGAR models if not stated otherwise.
}
\vspace{-.2cm}
\label{tab:roborefit_results}
\begin{tabular}{lcccc} \toprule
\multirow{2}{*}{Method} & \multicolumn{2}{c}{testA} & \multicolumn{2}{c}{testB} \\
&  Acc@0.5 & mIoU & Acc@0.5 & mIoU \\ \midrule
RefTR (r50)~\cite{lu2023roborefit} & 84.22 & 81.16 & 54.12 & 52.98 \\
RefTR (r101)~\cite{lu2023roborefit} & 81.19 & 78.07 & 45.68 & 49.75 \\  \midrule
SUGAR (no pre-train, {\footnotesize $N_e$=768}) & 87.30 & 79.02 & 50.64 & 51.49 \\
SUGAR (no pre-train) & 87.56 & 81.31 & 55.62 & 57.02  \\
SUGAR (SN\_s) & 88.02 & 81.84 & 52.90 & 56.80 \\
SUGAR (SN\_m w/o grasp) & 88.66 & 81.42 & 59.70 & 59.76 \\
SUGAR (SN\_m )& 89.05 & 81.75 & 61.85 & 60.53 \\
SUGAR (Ens\_m) & \textbf{89.47} & \textbf{82.11} & \textbf{65.04} & \textbf{62.80} \\ \bottomrule
\end{tabular}
\vspace{-1.5em}
\end{table}

\noindent\textbf{Experimental results.}
Table~\ref{tab:ocidref_results} presents the results for the OCID-Ref dataset. 
For fair comparison with previous work~\cite{karamcheti2023voltron}, we fix the visual encoder and only finetune the decoder in SUGAR pre-trained on the ensembled multi-object dataset (Ens\_m).
Results show that SUGAR outperforms the state-of-the-art 2D visual representations.

\noindent
RoboRefit is a more challenging and realistic dataset.
We compare with state-of-the-art transformer models proposed in~\cite{lu2023roborefit} based on RGB-D images for fair comparison with SUGAR using colored point cloud input.
The results are presented in Table~\ref{tab:roborefit_results}.
First, we can see that the number of point cloud tokens matters a lot from the first two rows in the SUGAR variants. The task requires high resolution point cloud embeddings.
Our SUGAR architecture trained from scratch achieves comparable or better performance than the state-of-the-art methods.
Pre-training in single-object datasets - SUGAR (SN\_s), does not benefit the object grounding in cluttered scene on the more difficult testB split.
SUGAR (SN\_m) pre-trained on ShapeNet multi-object dataset improves the performance of SUGAR w/o pre-training by around 6\% on testB Acc@0.5.
Interestingly, we find that learning the object affordances is beneficial for referring expression grounding, see comparison to (SN\_m w/o grasping).
SUGAR (Ens\_m) pre-trained on a larger dataset achieves the best performance and
outperforms previous best Transformer (r50) model in~\cite{lu2023roborefit} by more than 10\% on the challenging testB split on Acc@0.5, demonstrating the generalization ability of our SUGAR representation.

\begin{table*}
\centering
\small
\tabcolsep=0.14cm
\caption{Success rates of multi-task policies on 10 tasks of RLBench simulator.}
\vspace{-.2cm}
\label{tab:rlbench_10tasks}
\begin{tabular}{ccccccccccccc} \toprule
Method & Pre-train & \cellcolor{gray!15} Avg. & \begin{tabular}[c]{@{}c@{}}Pick \&\\ Lift\end{tabular} & \begin{tabular}[c]{@{}c@{}}Pick-Up\\ Cup\end{tabular} & \begin{tabular}[c]{@{}c@{}}Push\\ Button\end{tabular} & \begin{tabular}[c]{@{}c@{}}Put \\ Knife\end{tabular} & \begin{tabular}[c]{@{}c@{}}Put\\ Money\end{tabular} & \begin{tabular}[c]{@{}c@{}}Reach\\ Target\end{tabular} & \begin{tabular}[c]{@{}c@{}}Slide\\ Block\end{tabular} & \begin{tabular}[c]{@{}c@{}}Stack\\ Wine\end{tabular} & \begin{tabular}[c]{@{}c@{}}Take\\ Money\end{tabular} & \begin{tabular}[c]{@{}c@{}}Take\\ Umbrella\end{tabular} \\ \midrule
Auto-$\lambda$~\cite{guhur2023hiveformer} & - & \cellcolor{gray!15}69.3 & 87 & 78 & 95 & 31 & 62 & \textbf{100} & 77 & 19 & 64 & 80 \\
Hiveformer~\cite{guhur2023hiveformer} & - & \cellcolor{gray!15}83.3 & 88.9 & 92.9 & \textbf{100} & 75.3 & 58.2 & \textbf{100} & 78.7 & 71.2 & 79.1 & 89.2 \\ 
Hiveformer& R3M~\cite{nair2023r3m} & \cellcolor{gray!15}88.5 & 89.6 & 87.0 & 79.2 & 82.6 & 87.0 & \textbf{100} & 91.6 & 90.2 & 81.8 & 95.6 \\
Hiveformer & CLIP~\cite{radford2021clip} & \cellcolor{gray!15}87.2 & 87.2 & 96.0 & 68.8  & 68.8 & 94.8 & \textbf{100} & 92.4 & 93.0 & 73.2 & 97.6 \\ 
PolarNet~\cite{chen2023polarnet} & ShapeNetPart & \cellcolor{gray!15}89.8 & \textbf{97.8} & 86.0 & 99.6 & 80.5 & 94.1 & \textbf{100} & 93.4 & 80.5 & 68.1 & 97.8 \\ \midrule
\multirow{6}{*}{SUGAR} & - & \cellcolor{gray!15}85.9 & 77.7 & 92.7 & 91.7 & 69.4 & 87.7 & 99.7 & 94.3 & 83.1 & 66.8 & 95.7 \\
 & SN\_s & \cellcolor{gray!15}88.1 & 95.4 & \textbf{96.2} & 97.6 & 61.6 & 81.2 & \textbf{100} & 92.0 & 94.0 & 66.0 & 97.0 \\
 & SN\_m w/o grasp & \cellcolor{gray!15}91.9 & 94.9 & 94.1 & 90.9 & 83.9 & 92.5 & \textbf{100} & 97.0 & \textbf{96.2} & \textbf{72.3} & 97.0 \\
& SN\_m & \cellcolor{gray!15}\textbf{93.0} & 93.1 & 94.5 & 98.9 & 85.4 & \textbf{97.8} & \textbf{100} & \textbf{97.9} & 94.5 & 70.0 & 98.4 \\
& Ens\_m w/o grasp & \cellcolor{gray!15}92.0 & 93.1 & 93.7 & 98.8 & 85.5 & 92.3 & 99.9 & 97.3 & 93.7 & 68.8 & 97.2 \\
& Ens\_m & \cellcolor{gray!15}\textbf{93.0} & 95.8 & 95.7 & 96.1 & \textbf{86.5} & 94.2 & \textbf{100} & 97.0 & 93.5 & 72.0 & \textbf{98.8} \\
\bottomrule
\end{tabular}
\vspace{-.2cm}
\end{table*}

\subsection{Language-guided Robotic Manipulation}

This task aims to train a policy that can follow natural language instruction to perform manipulation tasks.
Our evaluation in this section is focused on the performance of multi-task policies and sample-efficient learning.

\noindent\textbf{Datasets.}
We evaluate models on the 10-task benchmark in the RLBench~\cite{james2020rlbench} simulator following previous work~\cite{liu2022autolambda,guhur2023hiveformer,chen2023polarnet}. The task names are listed in Table~\ref{tab:rlbench_10tasks}.
For each task, we use 100 demonstrations for behavior cloning.
Each demonstration consists of a sequence of keysteps of RGB-D image observations from three cameras and a 7-DoF action denoting the position, rotation and openness state of the gripper. The policy is required to predict keystep actions which are then executed by a motion planner. 
More details are provided in the supplementary material.

\noindent\textbf{Evaluation metric.}
The policy is evaluated by success rate. An episode achieves a score of 0 or 1 without any partial credits.
We perform three runs and report the average~\cite{guhur2023hiveformer,chen2023polarnet}.

\noindent\textbf{Downstream adaptation.}
We feed different prompt tokens into the decoder of SUGAR 
for action prediction, namely a masked current action token \verb|[act]|, a previous action token \verb|[pact]|, a step id \verb|[step]| and the language textual tokens. The output embeddings together with point embeddings from the encoder are used to predict actions via fully connected layers. More details are in the supplementary material.

\begin{figure}
    \centering
    \includegraphics[width=.9\linewidth]{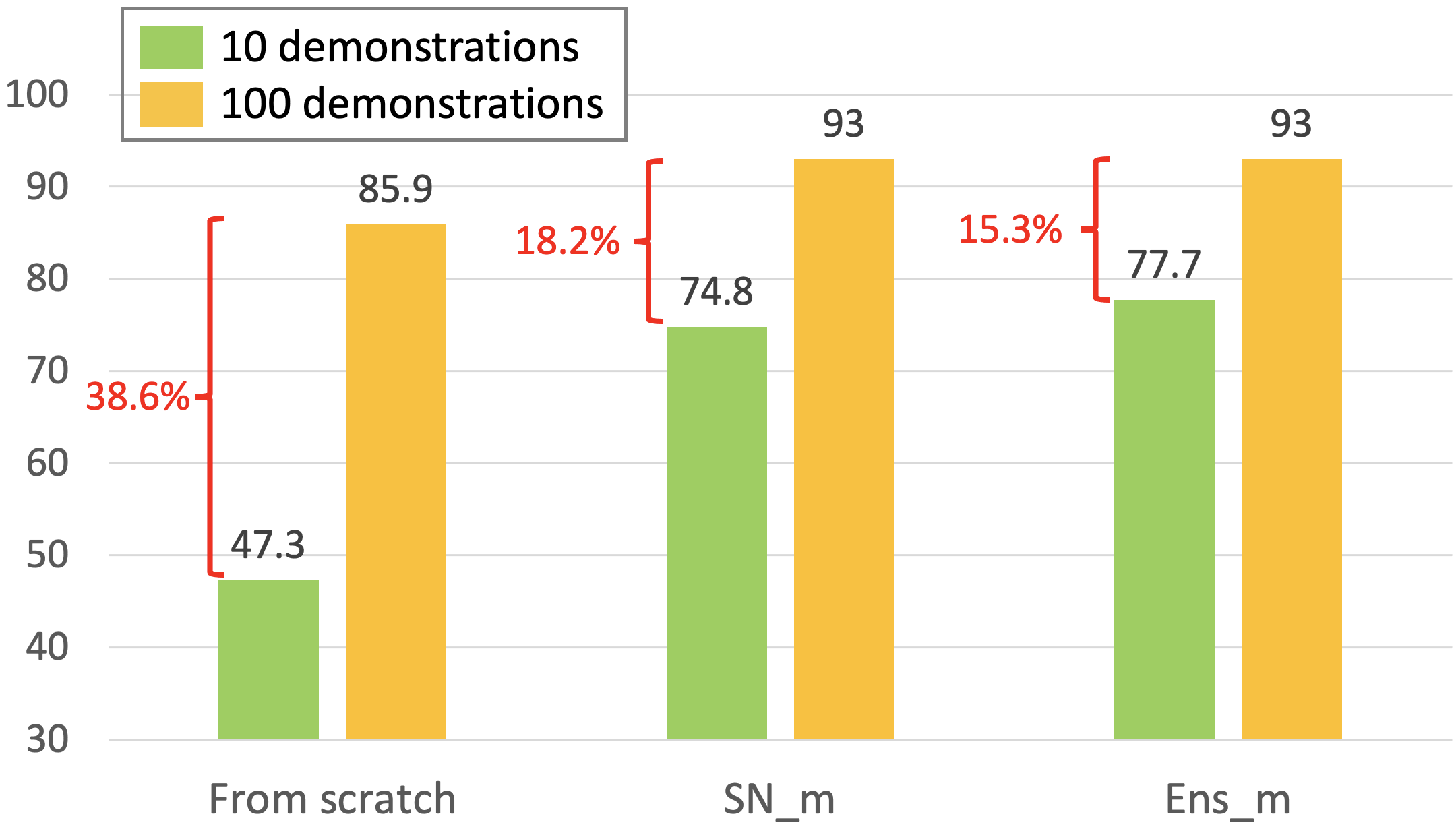}
    \vspace{-.2cm}
    \caption{Performance of training with 10 demonstrations.}
    \label{fig:rlbench_10demos}
    \vspace{-.4cm}
\end{figure}

\noindent\textbf{Experimental results.}
Table~\ref{tab:rlbench_10tasks} shows the results of multi-task policies on the 10 tasks.
We compare with the state-of-the-art methods and also improve the 2D-based method Hiveformer~\cite{guhur2023hiveformer} with pre-trained image backbones including R3M~\cite{nair2023r3m} and CLIP~\cite{radford2021clip}.
The image backbone is also finetuned end-to-end for action prediction otherwise the model performs poorly due to the large visual domain gap. 
Our model based on pre-trained SUGAR outperforms the 2D pre-trained models, PolarNet, and 3D models without pre-training. 
We also observe performance improvement with grasping prediction and multi-object scene in pre-training.
As there are sufficient training data for the policy, we do not observe further improvement using SUGAR (Ens\_m).
However, when we reduce the training data to 10 demonstrations per task, there is a clear advantage of Ens\_m representation compared to the SN\_m as shown in Figure~\ref{fig:rlbench_10demos}.
SUGAR (Ens\_m) significantly boosts the performance of the model trained from scratch with over 30\% improvement.
We further provide results on a real robot in the supplementary material and show the improvement from the pre-trained 3D representation for robot learning in real world.

\section{Conclusion}
\label{sec:conclusion}

This work presents SUGAR, a novel 3D pre-training framework for robotics.
It employs a versatile transformer-based architecture that jointly supports five pre-training tasks to learn semantic, geometric and affordances properties of objects in cluttered scenes.
Experimental results demonstrate the excellent performance when using SUGAR for three robotic-related tasks, namely, zero-shot 3D object recognition, referring expression grounding, and language-driven robotic manipulation.
Our work emphasizes the importance of cluttered scenes and object affordances when pretraining 3D representations for robotic applications.

{
\scriptsize
\noindent \textbf{Acknowledgements.} This work was partially supported by
the HPC resources from GENCI-IDRIS (Grant 20XX-AD011012122).
It was funded in part by the French government under management of Agence Nationale de la Recherche as part of the “Investissements d’avenir” program, reference ANR19-P3IA-0001 (PRAIRIE 3IA Institute), the ANR project VideoPredict (ANR-21-FAI1-0002-01) and by Louis Vuitton ENS Chair on Artificial Intelligence.
}

\ifarxiv \clearpage \appendix In Section~\ref{sec:impl_details}, we provide more implementation details for SUGAR pre-training and downstream adaptation.
Then in Section~\ref{sec:add_results} we present additional quantitative results.
We further perform real robot experiments in Section~\ref{sec:real_robot_expr} to demonstrate the effectiveness of SUGAR pre-training for robotic manipulation in the real world.
Finally, we discuss limitations and future work in Section~\ref{sec:limitation}.

\section{Implementation Details}
\label{sec:impl_details}

\subsection{Pre-training}

\nbf{Network details}
We set the number of points $N=4096$, the number of key points $N_e=256$ and the group size $S_e=32$ to obtain the point cloud input tokens.
The SUGAR encoder and decoder contains $L=12$ transformer blocks with hidden size $d=384$ and 6 attention heads per block.

\nbf{Training details}
We pre-train two sets of models according to the pre-training data: `SN' uses objects only in ShapeNet, and `Ens' uses the ensembled four datasets.
For the `SN' model, we train 100K iterations on the single-object dataset with learning rate 1e-4 and 100K iterations on the multi-object dataset with learning rate 1e-5 and batch size 128.
For the `Ens' model, we train 300K iterations on the single-object dataset and 200K iterations on the multi-object dataset using the same learning rate and batch size as in `SN' models.
The pre-training is performed on one NVIDIA-A100 GPU, taking 50 hours for the `SN' model and 130 hours for the `Ens' model.

\subsection{Referring expression grounding}
For the OCID-Ref dataset, we fix the point cloud encoder and only finetune the prompt-based decoder. We finetune the model with a batch size of 64 and learning rate of 1e-4 for 20 epochs.
For the RoboRefit dataset, we finetune the full model with a batch size of 16 and learning rate of 4e-5 for 50 epochs.
We use the AdamW optimizer with cosine learning rate scheduler.

\subsection{Language-guided robotic manipulation}

\nbf{Experimental setup}
Our experimental setup on RLBench~\cite{james2020rlbench} 10 tasks is the same as previous works~\cite{guhur2023hiveformer,chen2023polarnet}.
Specifically, we use three cameras located on the left shoulder, right shoulder and wrist of the robot with known camera intrinsics and extrinsics. Each camera produces an RGB-D image with image resolution of 128 $\times$ 128 at every step.
A merged point cloud can be obtained given the camera parameters.
Following~\cite{chen2023polarnet}, we only keep points inside the robot's workspace by using a fixed bounding box around the table. We use voxel downsampling to uniformly downsample the point cloud with 0.5cm grid size. 
For robotic control, we use keysteps~\cite{liu2022autolambda,guhur2023hiveformer,chen2023polarnet} - key turning points in action trajectories where the gripper changes its openness state or velocities of joints are close to zero.
The control policy should predict a position (3D), rotation (4D represented by quaternion) and openness state (1D) of the gripper for the next keystep.
The default motion planner in RLBench is used to find a trajectory between two keysteps. 

\nbf{Model details}
Figure~\ref{fig:rlbench_policy_architecture} illustrates the policy network in detail.
We first combine the action prompt embedding $y^L_1$ and point embeddings $\{x_i^L\}_{i=1}^{N_e}$ to compute a heatmap over all the key points, which denotes the importance of the key points for action prediction.
We then average the point embeddings and position of key points respectively using the heatmap. 
The averaged point embedding is concatenated with $y^L_1$ to regress the position offset relative to the averaged key point position, a rotation vector and an openness state.
The policy is trained by behavior cloning, with MSE loss for position and rotation, and BCE loss for openness state.

\nbf{Training details}
We use a batch size of 8 to train the model for 200K iterations for the 10 RLBench tasks.
We adopt a learning rate of 5e-5 for the model trained from scratch, while a lower learning rate of 2e-5 for the model initialized from SUGAR pre-training.

\begin{figure}
    \centering
    \includegraphics[width=1\linewidth]{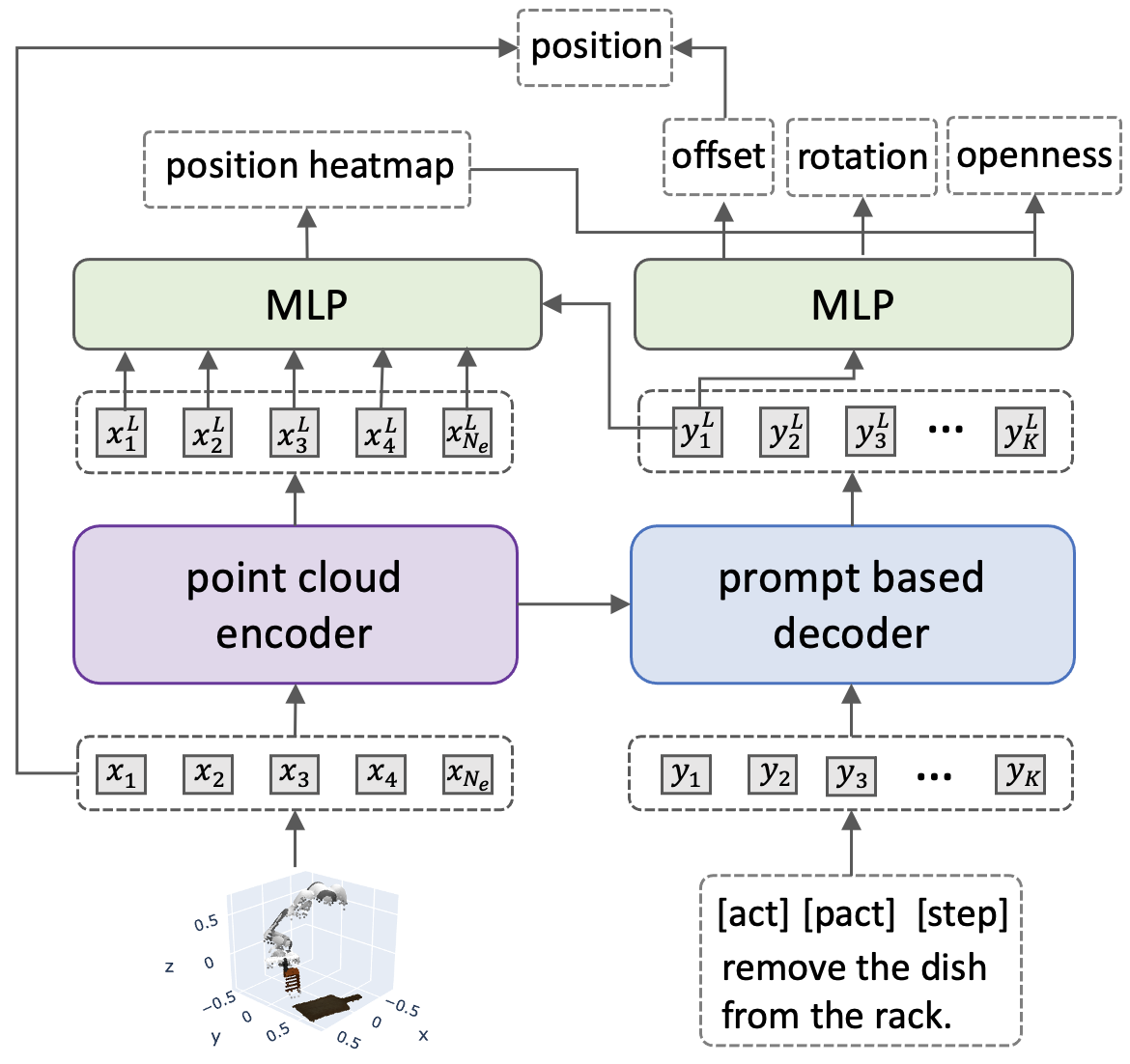}
    \caption{Network architecture for language-guided robotic manipulation. The point cloud encoder and prompt based decoder can be finetuned from SUGAR pre-training. We use two multi-layer perceptrons modules (MLP) as the action prediction head.}
    \label{fig:rlbench_policy_architecture}
\end{figure}

\section{Additional Results}
\label{sec:add_results}

\nbf{Zero-shot object recognition}
Though we consider the Ensembled w/o LVIS setup to be better suited for evaluating the generalization ability of models, we include results with LVIS training in Table~\ref{tab:rebuttal_zeroshot_eval} for complete comparison with prior work~\cite{liu2023openshape}. Training with LVIS split improves performance on the LVIS dataset but does not impact much on the other two datasets. Our model still outperforms the SoTA method~\cite{liu2023openshape} under this setup.

\begin{table*}[t]
\centering
\small
\caption{Zero-shot object recognition performance with models trained on Ensembled w/ LVIS dataset.}
\label{tab:rebuttal_zeroshot_eval}
\begin{tabular}{lccccccc} \toprule
\multirow{2}{*}{Method} & \multirow{2}{*}{ModelNet40} & \multicolumn{3}{c}{ScanObjectNN} & \multicolumn{3}{c}{Objaverse-LVIS} \\
 &  & OBJ\_ONLY & OBJ\_BG & PB\_T50\_RS & Top1 & Top3 & Top5 \\ \midrule
OpenShape~\cite{liu2023openshape} & 84.4 & 54.0 & 59.1 & 43.6 & 46.8 & 69.1 & 77.0 \\ \midrule
SUGAR (single) & \textbf{84.6} & \textbf{65.3} & \textbf{67.6} & \textbf{49.8} & \textbf{49.5} & \textbf{72.2} & \textbf{78.8} \\
SUGAR (multi) & 84.5 & 64.9 & 66.8 & 48.3 & 46.8 & 69.7 & 76.6 \\ \bottomrule
\end{tabular}
\end{table*}

\nbf{Referring expression grounding}
We provide an additional variant SUGAR (Ens\_s) in Table~\ref{tab:roborefit_results_suppmat}, which is pre-trained on single objects of the ensembled dataset.
To be noted, we only initialize the point cloud encoder for SUGAR variants pre-trained on single objects as we find initializing both encoder and decoder deteriorates the performance. As the decoder in single-object pre-training focuses on the overall scene for cross-modal learning, we hypothesize that the learned cross-modal attentions can suffer from recognition of local objects.
As shown in Table~\ref{tab:roborefit_results_suppmat}, the single object pre-training on the ensembled dataset does not benefit the generalization on unseen cluttered scenes in testB split, demonstrating the importance of pre-traning on multi-object scenes.

\begin{table}
\centering
\small
\tabcolsep=0.07cm
\caption{Performance of referring expression detection (evaluated by Acc@0.5) and referring expression segmentation (evaluated by mIoU) on the RoboRefit dataset.}
\label{tab:roborefit_results_suppmat}
\begin{tabular}{lcccc} \toprule
\multirow{2}{*}{Method} & \multicolumn{2}{c}{testA} & \multicolumn{2}{c}{testB} \\
&  Acc@0.5 & mIoU & Acc@0.5 & mIoU \\ \midrule
SUGAR (no pre-train) & 87.56 & 81.31 & 55.62 & 57.02  \\
SUGAR (Ens\_s) & 88.11 & 81.71 & 52.59 & 56.57 \\
SUGAR (Ens\_m) & \textbf{89.47} & \textbf{82.11} & \textbf{65.04} & \textbf{62.80} \\
\bottomrule
\end{tabular}
\end{table}

\begin{table*}[t]
\centering
\footnotesize
\tabcolsep=0.09cm
\caption{Averaged success rate of three runs for multi-task policies on 10 tasks of RLBench simulator.}
\label{tab:rebuttal_rlbench_10tasks}
\begin{tabular}{lcccccccccccc} \toprule
Method & Pre-train & Avg. & \begin{tabular}[c]{@{}c@{}}Pick \&\\ Lift\end{tabular} & \begin{tabular}[c]{@{}c@{}}Pick-Up\\ Cup\end{tabular} & \begin{tabular}[c]{@{}c@{}}Push\\ Button\end{tabular} & \begin{tabular}[c]{@{}c@{}}Put\\ Knife\end{tabular} & \begin{tabular}[c]{@{}c@{}}Put\\ Money\end{tabular} & \begin{tabular}[c]{@{}c@{}}Reach\\ Target\end{tabular} & \begin{tabular}[c]{@{}c@{}}Slide\\ Block\end{tabular} & \begin{tabular}[c]{@{}c@{}}Stack\\ Wine\end{tabular} & \begin{tabular}[c]{@{}c@{}}Take\\ Money\end{tabular} & \begin{tabular}[c]{@{}c@{}}Take\\ Umbrella\end{tabular} \\ \midrule
PolarNet~\cite{chen2023polarnet} & ShapeNetPart & $89.8_{\pm 1.5}$ & $97.8_{\pm 1.4}$ & $86.0_{\pm 2.1}$ & $99.6_{\pm 0.4}$ & $80.5_{\pm 1.1}$ & $94.1_{\pm 0.8}$ & $100_{\pm 0.0}$ & $93.4_{\pm 0.9}$ & $80.5_{\pm 3.6}$ & $68.1_{\pm 4.3}$ & $97.8_{\pm 0.2}$ \\ \midrule
\multirow{4}{*}{SUGAR} & - & $85.9_{\pm 3.9}$ & $77.7_{\pm 4.9}$ & $92.7_{\pm 4.2}$ & $91.7_{\pm 0.9}$ & $69.4_{\pm 8.0}$ & $87.7_{\pm 1.2}$ & $99.7_{\pm 0.4}$ & $94.3_{\pm 0.4}$ & $83.1_{\pm 7.8}$ & $66.8_{\pm 9.2}$ & $95.7_{\pm 1.6}$ \\
 & SN\_m & $93.0_{\pm 1.0}$ & $93.1_{\pm 1.3}$ & $94.5_{\pm 1.0}$ & $98.9_{\pm 0.8}$ & $85.4_{\pm 1.4}$ & $97.8_{\pm 1.3}$ & $100_{\pm 0.0}$ & $97.9_{\pm 0.8}$ & $94.5_{\pm 1.5}$ & $70.0_{\pm 1.6}$ & $98.4_{\pm 0.2}$ \\
 & Ens\_m w/o grasp & $92.0_{\pm 1.6}$ & $93.1_{\pm 1.3}$ & $93.7_{\pm 1.3}$ & $98.8_{\pm 1.1}$ & $85.5_{\pm 0.1}$ & $92.3_{\pm 5.3}$ & $99.9_{\pm 0.1}$ & $97.3_{\pm 1.4}$ & $93.7_{\pm 0.6}$ & $68.8_{\pm 4.2}$ & $97.2_{\pm 0.9}$  \\
 & Ens\_m & $93.0_{\pm 1.7}$ & $95.8_{\pm 1.3}$ & $95.7_{\pm 1.6}$ & $96.1_{\pm 5.1}$ & $86.5_{\pm 2.7}$ & $94.2_{\pm 1.6}$ & $100_{\pm 0.0}$ & $97.0_{\pm 0.5}$ & $93.5_{\pm 0.6}$ & $72.0_{\pm 2.9}$ & $98.8_{\pm 0.9}$  \\ \bottomrule
\end{tabular}
\end{table*}

\nbf{Language-guided robotic manipulation}
In Table~\ref{tab:rebuttal_rlbench_10tasks}, we include both the averaged success rate and standard deviations for the RLBench 10-task experiment.
As the 10 RLBench tasks use objects with simple shapes like cups and cubes, pre-training on ShapeNet can be sufficient and thus we do not observe further performance improvement from pre-training on Ens\_m.
Compared to PolarNet, our model performs slightly worse on Pick \& Lift and Push Button though it achieves better performance on average. To be noted, PolarNet employs additional normal and height features in the point cloud, while our method omits those for generalizability in pre-training.
As shown in PolarNet, normal and height features benefit some tasks like ``Push Button" where the main failure cases are that the gripper does not push down enough to the button.
We also notice relatively large variations on individual tasks, and thus we consider the averaged performance is more stable for comparison.

In Figure~\ref{fig:rlbench_10tasks_val}, we present the performance on the RLBench validation split for policies trained from scratch and initialized from SUGAR pre-training.
We can see that the policy can converge much faster and achieve better performance with the pre-training.

\begin{figure}
    \centering
    \includegraphics[width=0.9\linewidth]{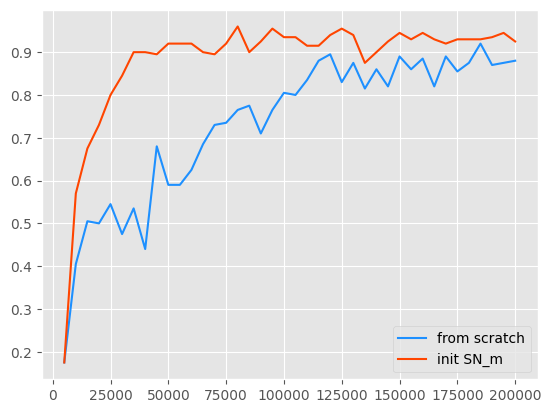}
    \caption{Success rate on RLBench validation split in different training iterations. We compare the policy trained from scratch and the model initialized from SUGAR pre-training.}
    \label{fig:rlbench_10tasks_val}
\end{figure}

\section{Real-world Robotic Manipulation}
\label{sec:real_robot_expr}

\begin{figure*}
     \centering
     \begin{subfigure}[b]{0.19\linewidth}
         \centering
         \includegraphics[width=\textwidth]{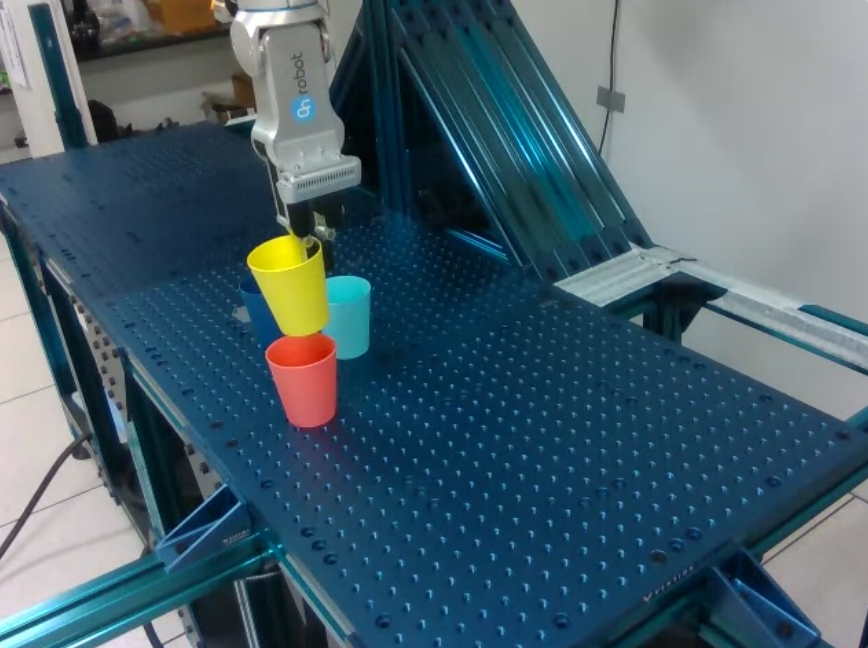}
         \caption{Stack cup.}
         \label{fig:real_task_stack_cups}
     \end{subfigure}
     \hfill
     \begin{subfigure}[b]{0.19\linewidth}
         \centering
         \includegraphics[width=\textwidth]{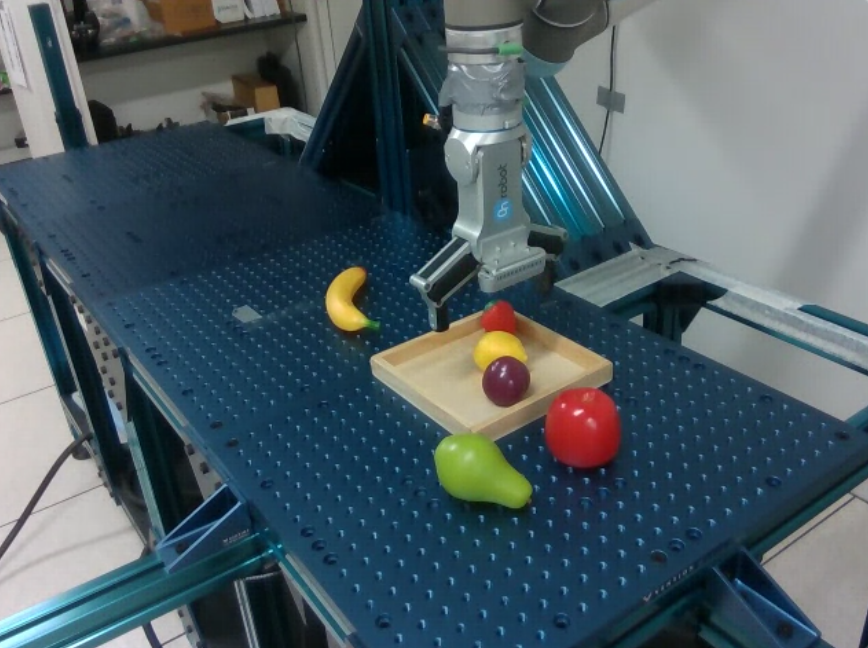}
         \caption{Put fruit in box.}
         \label{fig:real_task_put_fruit}
     \end{subfigure}
     \hfill
     \begin{subfigure}[b]{0.19\linewidth}
         \centering
         \includegraphics[width=\textwidth]{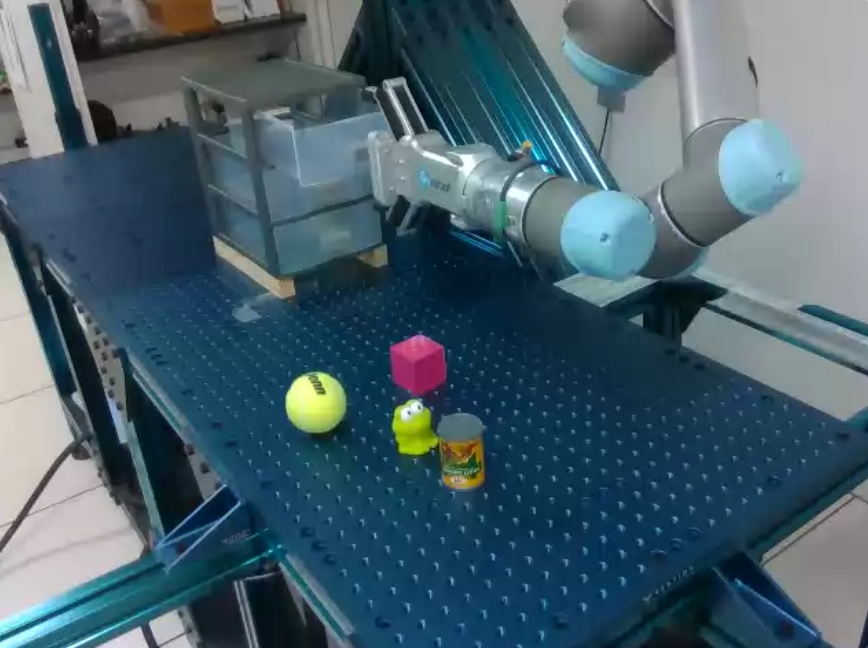}
         \caption{Open drawer.}
         \label{fig:real_task_open_drawer}
     \end{subfigure}
     \hfill
     \begin{subfigure}[b]{0.19\linewidth}
         \centering
         \includegraphics[width=\textwidth]{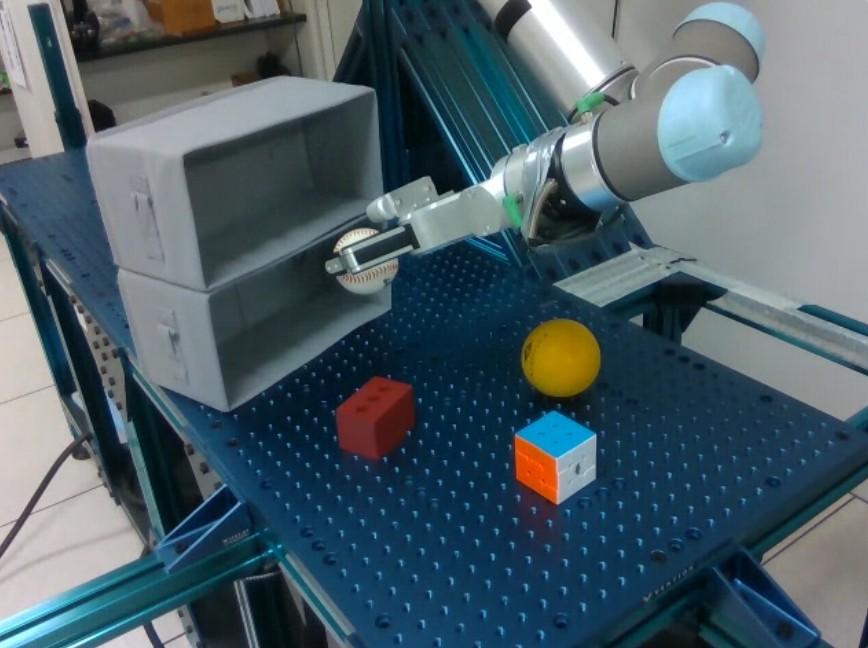}
         \caption{Put item in cabinet.}
         \label{fig:real_task_put_cabinet}
     \end{subfigure}
     \hfill
     \begin{subfigure}[b]{0.19\linewidth}
         \centering
         \includegraphics[width=\textwidth]{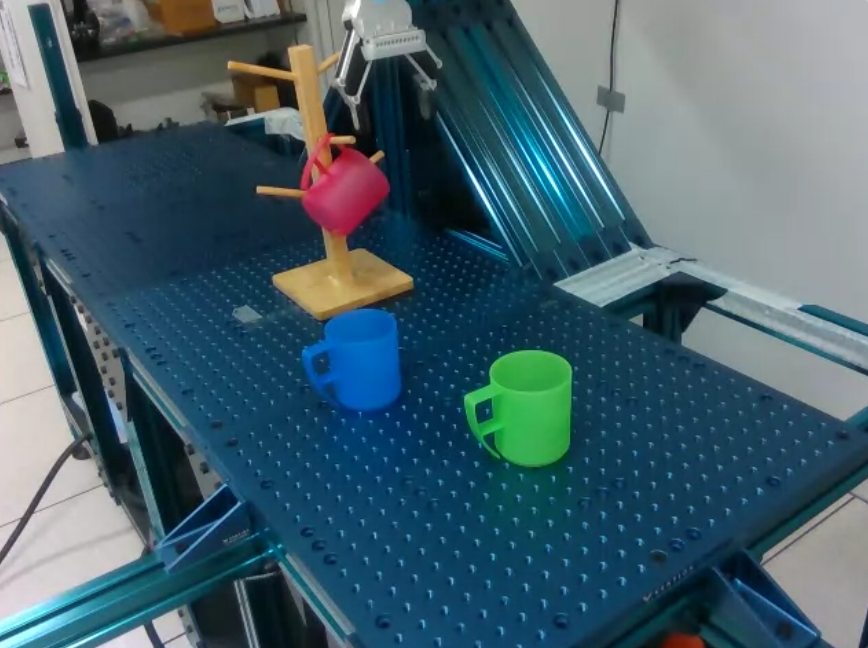}
         \caption{Hang mug.}
         \label{fig:real_task_hang_mug}
     \end{subfigure}
     
        \caption{Illustration of the adopted five real robot tasks.}
        \label{fig:real_robot_tasks}
\end{figure*}

To evaluate the effectiveness of SUGAR pre-training for real robots, we further perform real world experiments for language-guided robotic manipulation.

To be specific, we use a UR5 robotic arm equipped with a RG6 gripper and set two Intel RealSense D435 RGB-D cameras on the front and lateral sides of the robot's workspace.
We adopt 5 real-world tasks including \emph{stack cup}, \emph{put fruit in box}, \emph{open drawer}, \emph{put item in cabinet} and \emph{hang mug} as illustrated in Figure~\ref{fig:real_robot_tasks}.
For each task, we collect 20 real-robot demonstrations, where each demonstration consists of RGB-D images and proprioceptive information of the gripper at keysteps (typically 3-6 keysteps).

\begin{table}
\centering
\small
\caption{Success rate of multi-task policies on 5 real-world tasks. We evaluate 10 episodes for each task.}
\label{tab:real_world_exprs}
\begin{tabular}{lcc} \toprule
 & no pretrain & SUGAR (Ens\_m)  \\ \midrule
Stack cup & 0/10 & 10/10 \\
Put fruit in box & 0/10 & 4/10 \\
Open drawer & 0/10 & 3/10 \\
Put item in cabinet & 0/10 & 9/10 \\
Hang mug & 0/10 & 6/10 \\ \bottomrule
\end{tabular}
\end{table}

We train a multi-task policy using the collected real-robot data, and evaluate 10 episodes for each task where the object locations and distractor objects are different from the training data.
Table~\ref{tab:real_world_exprs} presents results of a model trained from scratch on the real robot data and a model initialized from SUGAR pre-training.
The model trained from scratch overfits on the limited training data and totally fails in evaluation.
As shown in Figure~\ref{fig:real_scratch_put_fruit_failure}, the model trained from scratch has serious problems of localizing the target object.
Our SUGAR pre-training significantly improves the performance for language-guided manipulation in the real world, leading to an average of 64\% success rate over the five tasks.
Figure~\ref{fig:real_sugar_put_fruit_success} presents a successful case of putting lemon in the box.
However, we also notice that the model initialized from SUGAR pre-training still has problems in precise object localization in Figure~\ref{fig:real_sugar_put_fruit_failure}.
The problems can result from the sub-optimal network design that largely downsamples the point cloud, the regression action prediction head that is more unstable compared to classification, and the noisy depth sensors.
We will investigate more on the policy networks to improve the robotic manipulation performance.

\begin{figure*}[t]
    \centering
     \begin{subfigure}[b]{1\linewidth}
         \centering
         \includegraphics[width=\linewidth]{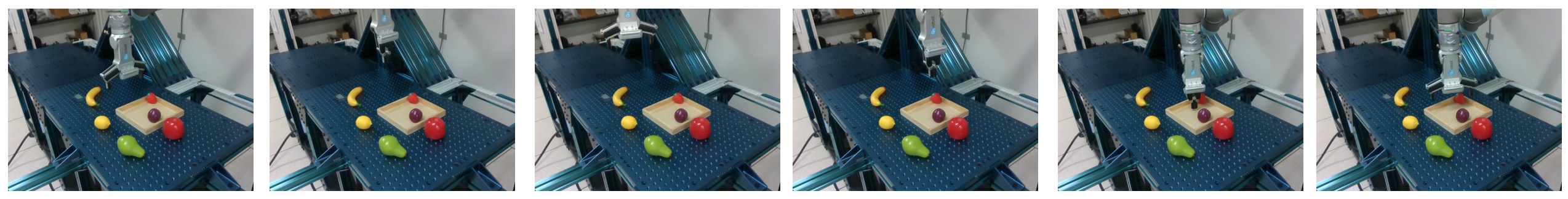}
         \caption{A failure case of  the multi-task policy trained from scratch.}
         \label{fig:real_scratch_put_fruit_failure}
     \end{subfigure}
     \hfill
     \begin{subfigure}[b]{1\linewidth}
         \centering
         \includegraphics[width=\linewidth]{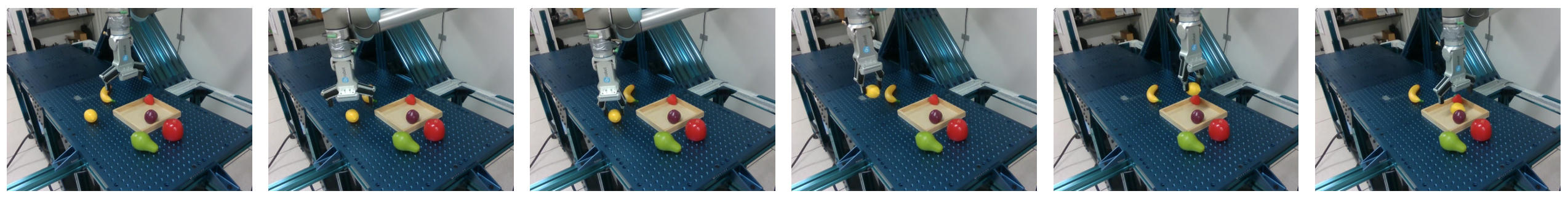}
         \caption{A successful case of  the multi-task policy initialized from SUGAR pre-training.}
         \label{fig:real_sugar_put_fruit_success}
     \end{subfigure}
     \hfill
     \begin{subfigure}[b]{1\linewidth}
         \centering
         \includegraphics[width=\linewidth]{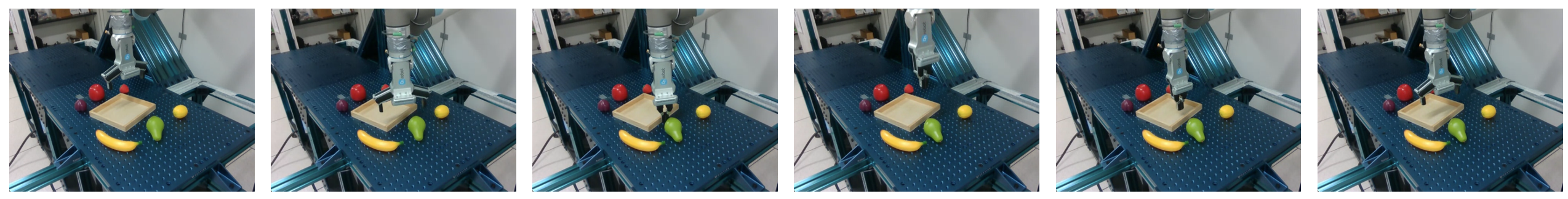}
         \caption{A failure case of  the multi-task policy initialized from SUGAR pre-training.}
         \label{fig:real_sugar_put_fruit_failure}
     \end{subfigure}
     \caption{Examples of real world execution on the \emph{Put fruit in box} task for different policies.}
        \label{fig:real_robot_examples}
\end{figure*}

\section{Limitations and Future Work}
\label{sec:limitation}

This work only adopt a plain transformer architecture for point cloud encoding, which is computationally expensive.
For example, compared to the SoTA method PolarNet~\cite{chen2023polarnet}, our model consists of 4.5x more parameters (65M vs. 14M) and runs 1.3x slower (18h vs. 14h in training on one V100 GPU).
This is because PolarNet is based on a UNet backbone which is more efficient.
Our vanilla transformer-based backbone alone does not a show clear advantage over the UNet backbone for robotic manipulation as seen in Table~\ref{tab:rebuttal_rlbench_10tasks}, although the proposed pre-training significantly boosts the performance.
We believe that the proposed pre-training can benefit other architectures and plan to explore more efficient 3D backbones in our future work.
 \fi

\end{document}